\documentclass[twocolumn,letterpaper,10pt]{article}
\usepackage[top=0.6in, bottom=1in, left=0.6in, right=0.6in]{geometry}
\usepackage[utf8]{inputenc}
\usepackage{amsmath, amssymb}

\usepackage{graphicx}
\usepackage{subfig}
\usepackage{authblk}
\usepackage{booktabs}
\usepackage{lipsum}

\title{An Investigation of Test-time Adaptation for Audio Classification under Background Noise}
\author[1]{Weichuang~Shao}
\author[2]{Iman~Yi Liao}
\author[3]{Tomas~Henrique~Bode~Maul}
\author[4]{Tissa~Chandesa}
\affil[1,2,3,4]{School of Computer Science, The University of Nottingham Malaysia, Semenyih, Malaysia \\ 
\texttt{andyshao90@gmail.com,\{Iman.Liao,Tomas.Maul,Tissa.Chandesa\}@nottingham.edu.my}}
\date{}

\begin{document}

\maketitle

\begin{abstract}
    Domain shift is a prominent problem in Deep Learning, causing a model pre-trained on a source dataset to suffer significant performance degradation on test datasets. This research aims to address the issue of audio classification under domain shift caused by background noise using Test-Time Adaptation (TTA), a technique that adapts a pre-trained model during testing using only unlabelled test data before making predictions. We adopt two common TTA methods, TTT and TENT, and a state-of-the-art method CoNMix, and investigate their respective performance on two popular audio classification datasets, AudioMNIST (AM) and SpeechCommands V1 (SC), against different types of background noise and noise severity levels. The experimental results reveal that our proposed modified version of CoNMix produced the highest classification accuracy under domain shift (5.31\% error rate under 10 dB exercise bike background noise and 12.75\% error rate under 3 dB running tap background noise for AM) compared to TTT and TENT. The literature search provided no evidence of similar works, thereby motivating the work reported here as the first study to leverage TTA techniques for audio classification under domain shift. The project code is available at Github~\footnote{https://github.com/Andy-Shao/TTA-in-AC.git}.
\end{abstract}

\vspace{1em}
\vspace{1em}

\section{Introduction}\label{sec:introduction}
Domain shift is a major factor that contributes to model performance degradation when a model pre-trained on a source dataset (e.g., clear speech signals) is deployed on target datasets (e.g., speech signals with background noise, different accents, and other variations) which have different distributions from the source dataset. Test-time Adaptation (TTA), a technique that has recently gained popularity, deals with the problem by adapting the pre-trained model, e.g., updating model parameters or model behavior, during test-time using only unlabeled test data~\cite{liang2024comprehensive}.

Depending on the manner through which test data is used to adapt a pre-trained model and, in turn, how the adapted model makes predictions, TTA can be categorized into Online Test-time Adaptation (OTTA), Test-time Batch Adaptation (TTBA), and Test-time Domain Adaptation (TTDA)~\cite{liang2024comprehensive}. OTTA assumes test data comes in sequentially, and the model is adapted for each sample and then used to predict the sample immediately. OTTA adaptation is accumulative over test samples. TTBA adapts the model for a batch of test data samples and predicts each batch's samples after adaptation. The model is reset to its pre-trained version after adaptation for each batch, so the adaptation for each batch is independent of all other adaptations. TTDA uses the entire test dataset to adapt the pre-trained model, and the same adapted model is used to make predictions for all the samples in the test dataset. OTTA, TTBA, and TTDA all follow the same TTA framework.

We are interested in how the above typical TTA methods may be applied to audio classification tasks under background noise and aim to answer the following research questions.
\begin{itemize}
    \item \emph{Q1}: Can typical TTA methods that are designed for Computer Vision tasks be adapted to audio classification tasks under background noise?
    \item \emph{Q2}: How do typical TTA methods, including TTT~\cite{sun2020test}, TENT~\cite{wang2021tent}, and CoNMix~\cite{kumar2023conmix}, perform on audio classification tasks with varying noise types and noise levels?
\end{itemize}

To this end, we adopt the AudioMNIST~\cite{audiomnist2023} (AM) and the SpeechCommands V1~\cite{warden2018speech} (SC) datasets to create audio classification tasks with background noise. The AM dataset contains samples of speech utterances of 10 digits from zero to nine. The SC dataset consists of samples of speech utterances of 30 different words including 10 digits too. Domain shift is simulated by adding background noise (e.g., runny tap) to the clean speech utterance in the testing samples. To address research question Q1, in order to directly use TTT~\cite{sun2020test}, TENT~\cite{wang2021tent}, and CoNMix~\cite{kumar2023conmix}, which were originally designed for image inputs, we adopt the popular Mel-Spectrogram~\cite{ustubioglu2023mel,hwang2020mel} to convert raw speech signals to 2D images based on the concept of Short Term Fourier Transform (STFT) (a.k.a., windowed Fourier Transform). Whenever augmentations are necessary, we replace image augmentation with augmentations that are suitable for time series, such as time shift. To address research question Q2, we analyze the performance of TTT~\cite{sun2020test}, TENT~\cite{wang2021tent}, and CoNMix~\cite{kumar2023conmix} against three different types of background noise, i.e., doing the dishes (DD), exercise bike (EB), and running tap (RT), each with two different noise levels, namely, 3.0 dB and 10.0 dB Signal-to-Noise Ratios (SNR).

Our investigation has revealed the following findings:
\begin{itemize}
    \item Literature search provided no evidence of similar works, we have, for the first time, reported negative adaptation effects by TTA methods for audio classification under background noise. Specifically, both TENT and TTT had significantly higher prediction error rates in the SC test set with all three types of background noise after test-time adaptation than without applying adaptation. 
    \item The more complex TTA method, CoNMix, performed consistently better than when no adaptation was applied, across different audio classification datasets with different types and levels of background noise, demonstrating the importance of combining multiple TTA strategies.
\end{itemize}

\section{Related Works}

\noindent The majority of research on TTA has been in the fields of Image Processing and Computer Vision~\cite{liang2024comprehensive}. TTA has also been applied to other fields, such as video processing~\cite{azimi2022self,chen2022source,zeyang2022unsupervised,wang2023test}, Natural Language Processing~\cite{wang2021efficient,shu2022test}, and Audio Analysis~\cite{ klejch2019lattice, kim2021test, kim2023sgem, lin2024continual, Amiri2024PathologySpeechDetection}. To elaborate,  \cite{klejch2019lattice} discusses audio analysis with TTA in Automatic Speech Recognition (ASR). Following the typical components of unsupervised adaptation for ASR (i.e., selection of a set of model parameters for adaptation, filtering test data that are suitable for adaptation, and a reliable adaptation schedule preventing overfitting to test data), in~\cite{klejch2019lattice} the authors focused on the filtering process and used the lattice (i.e., a graph structure showing competing predictions) generated by the unadapted model for the test data to determine when the test data was suitable for test-time adaptation. 
\cite{kim2023sgem} contributes a Sequential-level Generalized Entropy Minimization (SGEM) framework for ASR when performing TTA. \cite{kim2023sgem} requires the beam search-based logit acquisition to select the most plausible logit for adaptation. However, audio classification does not require beam search-based logit acquisition; therefore, a sequential-level framework in~\cite{kim2023sgem} is not suitable for audio classification. In~\cite{lin2024continual}, the author contributes a continual Fast-slow TTA (CTTA) for ASR, which is based on Single-Utterance Test-time Adaptation (SUTA)~\cite{lin2022listen} that focuses on a CTC-based end-to-end ASR model instead of audio classification.
In~\cite{kim2021test}, TTA was investigated for Personalized Speech Enhancement on edge devices or scenarios where computational resources are limited. It is premised on the assumption that a clean and large teacher model exists and can generalise well to unseen person's speeches for enhancement. A student model is then fine-tuned during test-time to enhance a person's speech by comparing its model output with that of the teacher model. In \cite{Amiri2024PathologySpeechDetection}, TTA was explored for pathological speech detection for speech disorders in noisy environments, however, it still assumes access to a subset of training data or validation data during the testing phase, which is not the case in a TTA setting. The aforementioned methods either make unrealistic assumptions for a TTA setting in audio analysis, or do not extend to audio classification tasks directly, for which TTA has not been investigated, to the best of our knowledge.

\subsection{Meta-Learning}
\noindent Meta-Learning consists of a set of tasks, where the tasks own their independent training dataset and validation dataset~\cite{hospedales2021meta}. After each meta-task trains on its training dataset independently, the final task utilizes the weights of meta-tasks and validation datasets for the final tuning~\cite{hospedales2021meta}. For example, MAML~\cite{finn2017model} leverages meta-tasks to speed up final task adaptation by a few unseen samples and gradients. TTBA leverages Meta-Learning during the test-time stage, such as with backward propagation (MLSR~\cite{park2020fast} and Full-OSHOT~\cite{borlino2022self}) and forward propagation~\cite{dubey2021adaptive,kim2022variational}.

\subsection{Transfer Learning}
\noindent Transfer Learning utilises a particular model structure with two or more tasks, either supervised or unsupervised~\cite{goodfellow2016deep,tan2018survey}. In Transfer Learning, the main supervised task and other auxiliary unsupervised or self-supervised tasks share a part of the weights. In the test stage, as the labels of the main task are not available, only unsupervised or self-supervised auxiliary tasks are processed for improving the prediction performance of the main supervised task, e.g., TTT~\cite{sun2020test} leverages Transfer Learning strategies when performing TTA, and TTT only processes a self-supervised task (predicting the angle of image rotation) in the test stage.


In contrast, TTA imposes fewer constraints on the usage of source datasets because it does not require the i.i.d. assumption. Although this requirement limits conventional Transfer Learning, \cite{zhang2022transfer} introduces Transfer Adaptation Learning, an approach combining Transfer Learning and Domain Adaptation, to handle scenarios that do not satisfy the i.i.d. assumption, such as TTT~\cite{sun2020test}.

\subsection{Domain Adaptation}
\noindent Domain Adaptation is a technique to adjust a trained model in the source domain (training set) to work effectively in a different target domain (test set)~\cite{farahani2021brief}. There are three common Domain Adaptation techniques including Supervised Domain Adaptation (SDA), Unsupervised Domain Adaptation (UDA), and Semi-supervised Domain Adaptation (SSDA). For SDA, both source and target domain labels are available, but the domain distribution is different between source and target domains. Thus, SDA aims to shift a trained model from the source to the target domain~\cite{farahani2021brief,motiian2017unified,koniusz2017domain}. SDA usually learns the target domain task solution with low risk~\cite{farahani2021brief}. UDA deals with the scenarios where the labels in a target domain are not available~\cite{farahani2021brief,motiian2017unified}, whereas SSDA assumes that a small number of labels are available in the target domain~\cite{yao2015semi,saito2019semi}. TTA is closely related to Domain Adaptation \cite{wang2021tent,zhang2021source}, and many TTA methods are inspired by domain adaptation theories, such as \cite{li2020model,zhang2021source,hou2021visualizing,li2021imbalanced,ye2022alleviating}.

\section{Methodology}\label{sec:methodology}
\noindent This research investigates and evaluates two OTTA methods, TENT~\cite{wang2021tent} and TTT~\cite{sun2020test}, and a TTDA method, CoNMix~\cite{kumar2023conmix}, for the problem of audio classification in the context of background noise. It is worth noting that even though TTT has a TTBA version, its adaptation performance is inferior to its OTTA version (most probably due to the fact that the TTBA strategy does not utilise the entire test set for model adaptation), and hence it is not included in our methodology. 

\subsection{Mathematical Notation}

\noindent Let $\mathcal{X_S, X_T} \subseteq \mathbb{R}^d$ be the features of the training and test sets, and $\mathcal{Y_S, Y_T} \subseteq \mathbb{R}^c$ the corresponding labels of the training set and pseudo-labels of the test set, respectively. Assume that there are $n$ samples in the training set, i.e., $|\mathcal{X_S}|=|\mathcal{Y_S}|=n$, and $m$ samples in the test set, i.e., $|\mathcal{X_T}|=|\mathcal{Y_T}|=m$. Note that $\mathcal{Y_S}$ and $\mathcal{Y_T}$ are $c$-cardinality label sets.

Assume that the classification model is $f: \mathbb{R}^d \rightarrow \mathbb{R}^c$.
\begin{equation}
    \hat{y} = \sigma(f(x;\theta)); \forall x \in \mathcal{X_S} \bigcup \mathcal{X_T}
\end{equation}
where $\hat{y} \in \{0,1\}^c$ is the prediction of the input vector $x \in \mathbb{R}^d$, $\theta$ is the weights of $f$, and $\sigma(\cdot): \mathbb{R}^c \rightarrow \mathbb{R}^c$ is activation function.

For a model $f(x;\theta)$ pre-trained on source dataset $\mathcal{X_S, Y_S}$, the objective of TTA is to update part of or the entire weights $\theta$ to $\theta'$ so that the model $f(x;\theta')$ can achieve competitive performance for $x \in \mathcal{X_T}$ compared to that of $f(x;\theta)$ for $x \in \mathcal{X_S}$.

\subsection{TTT}
\noindent TTT~\cite{sun2020test} is a Transfer Learning model that is composed of three parts of weights including $\theta_{sh}$, $\theta_{cl}$, and $\theta_{tf}$. Specifically, TTT contains two classification models, one for the main classification task $f(x;\theta_{sh}, \theta_{cl})$ and the other for predicting image rotation angle $f(x;\theta_{sh}, \theta_{tf})$. During pre-training, both tasks utilise a cross-entropy loss function 
\begin{equation}\label{Cross_Entropy}
    \mathcal{L}_{ce}(\mathcal{X},\mathcal{Y};\theta) = - \sum_{x \in \mathcal{X}} \bar{y} \cdot \log \sigma(f(x;\theta))
\end{equation}
where $\mathcal{X}$ is the feature set; $\mathcal{Y}$ is the true label set; $\bar{y} \in \mathcal{Y}$ which is the true label. The total loss is the combination of the cross-entropy loss of the two tasks 
\begin{equation}\label{TTT_Pre_train_Loss}
    \begin{split}
        \mathcal{L}_{total} & = \mathcal{L}_{ce}(\mathcal{X_S},\mathcal{Y_S};\theta_{sh}, \theta_{cl}) \\
        & + \mathcal{L}_{ce}(\mathcal{X_S},\bar{Y}_\mathcal{S};\theta_{sh}, \theta_{tf})
    \end{split}
\end{equation}
where $\bar{Y}_\mathcal{S}$ denotes the true label set of image rotation angles in the training set.

As for the Test-time Adaptation stage, TTT only leverages the loss function for the task of predicting image rotation angles to update $\theta_{sh}, \theta_{tf}$, which is
\begin{equation}\label{TTT_Adaptation_loss}
    \mathcal{L}_{ce}(\mathcal{X_T}, Y_\mathcal{T};\theta_{sh},\theta_{tf})
\end{equation}
where $Y_\mathcal{T}$ is the true label set of image rotation angles in the test set.

As the task of predicting spectrogram rotation angles are meaningless for audio datasets, we replace the task with predicting time shifts in the audio dataset. We design three types of time shifts, including no shift, left shift, and right shift. For example, if an audio sample has 16 kHz sampling rate and is 1 second long, its shape would be [1, 16000]. We then left or right shift the audio using PyTorch's roll method with a specified percentage.

\subsection{TENT}
\noindent TENT~\cite{wang2021tent} requires that a model have Batch Normalisation layers, and during test-time adaptation, only the batch normalisation statistics are updated. Hence, it is efficient and can respond quickly during inference. TENT has two different types of implementation for TTA, namely Tent Adaptation and Norm Adaptation. 

\emph{Tent Adaptation} requires a standard entropy as the loss function: 
    \begin{equation}\label{Entropy_Equation}
        \mathcal{L}_{EN} = - \sum_{x \in \mathcal{D_T}} \sigma(f(x;\theta)) \cdot \log{\sigma(f(x;\theta))}
    \end{equation}
During test-time adaptation, the entropy loss (\ref{Entropy_Equation}) will be backpropagated to calculate the gradient of the parameters in BatchNorm layers. However, the training set's mean and standard deviation will be abandoned during forward propagation.

\emph{Norm Adaptation} only involves the BatchNorm forward propagation, where the mean and the standard deviation shifts are evaluated as follows.
\begin{equation}\label{BatchNorm_Equation}
    BN(x) = \frac{x - \mathbb{E}[x]}{\sqrt{\mathbb{VAR}[x] + \epsilon}} \cdot \gamma + \beta; x \in \mathcal{X_T}
\end{equation}
where $\mathbb{E}[x]$ is the expectation of $\mathcal{X_T}$; $\mathbb{VAR}[x]$ is the variance of $\mathcal{X_T}$; $\epsilon$ is a tiny decimal, such as 1e-8, to avoid dividing by zero; $\gamma$ and $\beta$ are learnable parameters. Note that Norm Adaptation does not need backward propagation.

As TENT~\cite{wang2021tent} focuses on updating BatchNorm layers during test-time, we replace the original model in TENT with ResNet50 \cite{he2016deep}, a well-performing CNN architecture in many image classification tasks, which includes multiple BatchNorm layers. 

\subsection{CoNMix}
\noindent CoNMix~\cite{kumar2023conmix} is a complex TTA method that falls under the category of TTDA where the information of the entire test set in the target domain $\mathcal{X_T}$ is shared and used for model adaptation and inference during test time~\cite{nelakurthi2018source}. More specifically, CoNMix leverages the Vision Transformer~\cite{dosovitskiy2020image} architecture instead of CNNs to learn an image classification model for the purpose of domain adaptation at test time. CoNMix was trained and evaluated for TTA using the Office-Home dataset~\cite{venkateswara2017deep}. 

For the purpose of TTA, CoNMix adopts a typical teacher-student learning paradigm and involves three steps, i.e., Pre-training, Single Target Domain Adaptation (STDA), and Multi-target Domain Adaptation (MTDA).

\subsubsection{Pre-training} CoNMix trains the model as an ordinary classification task with cross-entropy loss using samples in only one of the four domains in the Office Home dataset.

\subsubsection{STDA}
In this stage, the pre-trained model is copied, including the model architecture and weights, to build a teacher model for each target domain as a result of TTA. To adapt the pre-trained model for a specific target domain, three different TTA strategies are combined in CoNMix including Nuclear-norm Maximisation as a variant of entropy minimisation of model predictions ($\mathcal{L}_{NM}$), cross-entropy minimisation between model predictions and pseudo-labels generated for the test set ($\mathcal{L}_{ce}^{pl}$), and inconsistency minimisation between model predictions for different augmentations of the same input ($\mathcal{L}_{cons}$), as follows.  
\begin{equation}\label{STDA_loss}
    \mathcal{L}_{STDA} = \lambda_1 \mathcal{L}_{NM} + \lambda_2 \mathcal{L}_{ce}^{pl} + \lambda_3 \mathcal{L}_{cons}
\end{equation}
where $\lambda_1,\lambda_2,\lambda_3 \in \mathbb{R}$ are scaling factors.

$\mathcal{L}_{NM}$ has been reported to perform better than standard entropy minimisation, where
\begin{equation}\label{Nuclear_norm_loss}
    \mathcal{L}_{NM} = - ||f(x;\theta)||_F
\end{equation}
with $||A||_F = \sqrt{\sum_{i=1}^I \sum_{j=1}^J A_{i,j}}$ being Frobenius norm of $A \in \mathbb{R}^{I \times J}$.

To generate pseudo labels for all the test samples, CoNMix first collects predicted labels for all the samples in a test set using the pre-trained model. The central feature point for each class is then obtained to determine the pseudo label for each test sample based on the maximum cosine similarity between the feature of the specific sample and that of each class centre. As the resultant pseudo labels are usually noisy, they are further refined through an iterative scheme. 
The final Pseudo-label loss is as follows:
\begin{equation}\label{CoNMix_pl_ce_loss}
    \mathcal{L}_{ce}^{pl} = - \underset{(x,y^n(x)) \in (\mathcal{X_T}, \mathcal{Y_T})}{\mathbb{E}} \sum_{k=1}^c 1_{[k = y^n(x)]} \log \sigma(f(x;\theta))
\end{equation}
where $y^n(x)$ is the final refined pseudo label for sample $x$.

$\mathcal{L}_{cons}$ measures the inconsistency between the predictions of a strong augmentation $x_s$ and a weak augmentation $x_w$ of the same input,
\begin{equation}\label{eq:L_cons}
    \mathcal{L}_{cons} = - \underset{\hat{y}_s \in \hat{Y}_s}{\mathbb{E}} \sum_{k = 1}^c \hat{y}_w^k \log \hat{y}_s^k
\end{equation}
where $\hat{y}_s, \hat{y}_w$ are predictions of $x_s, x_w$, respectively; $\hat{Y}_s$ is the prediction for the strong augmentation within each batch. 

\subsubsection{MTDA}
In this stage, a single student model is trained for multi-target domain adaptation. Each adapted model for a specific target domain in the STDA stage works as an independent teacher model here to generate pseudo labels for each test sample in its own target domain. Random samples from two different target domains are then linearly combined to generate synthetic samples to mimic the test set for multi-target domains. The student model, which has the same architecture as the pre-trained model, is then trained from scratch using the synthetic test set with the following loss function:
\begin{equation}
    \begin{split}
        \mathcal{L}_{MTDA} & = \lambda \cdot \text{CE}(\Tilde{y}_{ij}, \hat{y}_{ij}) \\ 
        & + (1-\lambda) \cdot \text{CE}(\Tilde{y}_{ij}, \hat{y}_{ij})
    \end{split}
\end{equation}
where $\text{CE}(y_1, y_2) = - y_2 \log{y_1}$ is the Cross-Entropy loss function; $\Tilde{y}_{ij}$ is the class label of a synthetic sample generated by linearly mixing the pseudo label of a sample from the target domain $i$ and the pseudo label of a sample from the target domain $j$; $\hat{y}_{ij}$ is the predicted label by the student model for the linear mix of the corresponding samples from target domains $i$ and $j$.

\subsection{Our implementation of CoNMix for audio datasets}
\noindent As for the activation function and loss function, the original CoNMix adopts the Softmax (Eq.\ref{softmax}) in $\mathcal{L}_{ce}^{pl}$ as follows:
\begin{align}
    \sigma(x)^i & = \frac{e^{x^i}}{\sum_{j=1}^c e^{x^j}} \\
    \text{softmax}(x) & = [\sigma(x)^0, \cdots, \sigma(x)^c]^{\top} \label{softmax}
\end{align}

However, during our experimentation, we observed that the contribution of the original pseudo-label loss function in CoNMix to the performance of TTA was inconsistent across different datasets. For datasets such as Office Home and AM, which exhibit semantic domain shift between source and target domains (e.g., from Art to Real World in Office Home, and from German to Non-German accent in AM), CoNMix with its original pseudo-label loss led to satisfactory adaptation of the model at test time. Whereas for the SC which does not involve semantic domain shifts among the original samples, when we split the dataset into training and testing and added background noise to the test set to mimic domain shifts, the original CoNMix method exhibited some negative adaptation effects, i.e., the prediction error rate was higher after the adaptation compared to using the pre-trained model without any adaptation. We performed an ablation study (see Appendix~\ref{app:ablation_study}) on the exclusion of each of the three loss functions in the original CoNMix method, and identified that the inclusion of pseudo label loss (Eq.\ref{CoNMix_pl_ce_loss}) was responsible for the negative adaptation effect on SC. Please refer to the Ablation Study in Appendix~\ref{app:ablation_study} for further details.

To address the issue, we looked into the pseudo label loss and modified the original CoNMix pseudo label loss function (Eq.\ref{CoNMix_pl_ce_loss}). Specifically, for the AM, we removed the pseudo label loss $\mathcal{L}_{ce}^{pl}$ for the adaptation to the target domain. For the SC, SCR, and SCN, we replaced the Softmax with the Log Softmax (Eq.\ref{log_softmax}) in CoNMix,
\begin{align}
    \sigma(x)^i & = \log \frac{e^{x^i}}{\sum_{j=1}^c e^{x^j}} \\
    \text{log softmax}(x) & = [\sigma(x)^0, \cdots, \sigma(x)^c]^{\top}\label{log_softmax}
\end{align}
and also replaced the pseudo label loss $\mathcal{L}_{ce}^{pl}$ with negative likelihood loss $\mathcal{L}_{nll}^{pl}$ as follows,
\begin{align}
    w^i & = \text{weight}[i]\\
    l^i & = - w^i y^i \bar{y}^i\\
    \mathcal{L}_{nll}^{pl} & = \sum_{k=1}^c \frac{1}{\sum_{i} w^i} l^k \label{CoNMix_pl_nll_loss}
\end{align}
where $i \in \{1,\dots,c\}$ is the $i^{th}$ class; $(\bar{y},y) \in (\bar{Y}_\mathcal{T}, \mathcal{Y_T})$; $(\cdot)^i$ is the $i^{th}$ dimension; $\text{weight}[\cdot]$ is the weight array of $c$ classes which are all ones in our implementation; $\mathcal{L}_{nll}^{pl}$ is the mean reduction negative likelihood loss.
The final loss function in the modified CoNMix for SC is as follows:
\begin{equation}\label{speech_commands_STDA_loss}
    \mathcal{L}_{STDA} = \lambda_1 \mathcal{L}_{NM} + \lambda_2 \mathcal{L}_{nll}^{pl} + \lambda_3 \mathcal{L}_{cons}
\end{equation}
where $\lambda_1, \lambda_2, \lambda_3 \in \mathbb{R}$ are scaling factors.


\section{Experiments}\label{sec:experiment}
\noindent In this section, we report the experimental results of TENT (including both Tent Adaptation and Norm Adaptation), TTT, original CoNMix, and modified CoNMix on AM and SC as discussed in the Methodology section.

\subsection{Datasets}

\begin{table}
    \centering
    \caption{Dataset Distribution}
    \label{tab:dataset_distribution}
    \begin{tabular}{cccl}
        \toprule
        Dataset & Train & Validation & Test  \\
        \midrule
        AM & 20000 & N/A & 10000 \\
        SC & 51088 & 6798 & 6835 \\
        SCR & 32185 & 3577 & 15326 \\
        SCN & 18620 & 2552 & 2494 \\
        \bottomrule
    \end{tabular}
\end{table}
\noindent Table~\ref{tab:dataset_distribution} summarises the training, validation, and test data splits, for the AudioMNIST~\cite{audiomnist2023} (AM) and SpeechCommands V1~\cite{warden2018speech} (SC) that we used in this research, whereby the sampling rates for AM and SC were 48 kHz and 16 kHz, respectively. For the AM dataset, the training set includes only German accents whilst the test set includes additional accents apart from German. For the SC dataset, we adopt the default published training/validation/testing split. We further create a SpeechCommands Random (SCR) dataset as a random subset of the SC dataset with a random split of training/testing data. Specifically, 30\%, 63\%, and 7\% of source domain data (the training set of SC) is used as test, training, and validation data, respectively. The last dataset is a subset of the SC dataset containing only speech utterances of the numbers 0-9 in English, which we name SpeechCommands Numbers (SCN).

\subsection{Dataset Pre-processing}
\noindent The data formats used in image and audio modalities are generally different. 
Specifically, an image usually comprises three channels (green, blue, and red) and has two spatial dimensions (x and y coordinates). Each pixel in a channel represents the intensity of the colour at that specific location in that specific channel. On the other hand, audio data can have multiple channels but commonly consists of one or two channels, and each channel is a waveform along the time dimension.

To apply TTA methods that are designed for processing images, we adopt the Mel-Spectrogram~\cite{ustubioglu2023mel,hwang2020mel} to convert an audio signal to a 2D image.

To simulate corrupted audio data at test time, we add background noise to the original uncorrupted audio data in the test set. Assume $\mathbf{x}$ is an original test set sample, $x$ is its corrupted version, and $\mathbf{n}$ is the noise. Then, the data corruption function $ds(\cdots)$ is as follows:
\begin{equation}\label{eq:addnoise}
    x = ds(\mathbf{x}, \mathbf{n}) = \mathbf{x} + \mathbf{n} \sqrt{\frac{||\mathbf{x}||_2^2}{||\mathbf{n}||_2^2} \times 10^{-\frac{\text{SNR}}{10}}}
\end{equation}
where SNR is the desired signal-to-noise ratio between $\mathbf{x}$ and $\mathbf{n}$. We design two noise levels, i.e., 3.0 dB and 10.0 dB, for three different types of background noise, which are covered by SC and include 'doing the dishes' (DD), 'exercise bike' (EB), and 'running tap' (RT).

As the length of an audio sample is one second in both AM and SC, whereas the length of the three types of background noise is more than one minute, we choose a random clip of one second in a source background noise clip to add to every test data sample according to Eq.\ref{eq:addnoise}.

Note that the same training and test sets are used for TENT, TTT, and CoNMix for their pre-training and test-time adaptation in our subsequent experiments.

\subsection{SNR at 10.0 dB Level}

\noindent Table~\ref{tab:resultsAll} includes the prediction error rates of all four TTA models across three datasets, including AM, SC, and SCR, adapted to different background noise types (DD, EB, and TR) at a 10 dB SNR level. 


\begin{table}[h]
    \centering
    \caption{TTA results of all models in terms of prediction error rates across three datasets (AM, SC, and SCR) for three types of background noise (DD, EB, and RT), at 10 dB and 3 dB SNR noise level, respectively.}
    \label{tab:resultsAll}
    \begin{tabular}{ccccccl}
        \toprule
        Dataset & Noise-SNR & Tent & Norm & TTT & CoNMix  \\
        \midrule
        AM & DD-10 & 17.23 & 19.34 & 53.85 & \textbf{6.27} \\
        SC & DD-10 & 77.32 & 82.74 & 9.17 & \textbf{8.49} \\
        SCR & DD-10 & 18.27 & 22.97 & 9.35 & \textbf{9.42} \\
        \midrule
        AM & DD-3 & 46.37 & 46.06 & 85.81 & \textbf{12.85} \\
        SC & DD-3 & 82.75 & 85.87 & 19.93 & \textbf{14.47} \\
        SCR & DD-3 & 33.52 & 39.66 & 19.87 & \textbf{16.29} \\
        \midrule
        AM & EB-10 & 12.44 & 15.63 & 51.46 & \textbf{5.31} \\
        SC & EB-10 & 79.2 & 83.88 & 10.86 & \textbf{8.32} \\
        SCR & EB-10 & 19.78 & 26.41 & 12.04 & \textbf{9.02} \\
        \midrule
        AM & EB-3 & 33.63 & 35.92 & 89.91 & \textbf{15.01} \\
        SC & EB-3 & 83.46 & 86.2 & 21.64 & \textbf{14.13} \\
        SC & EB-3 & 34.27 & 41.43 & 24.98 & \textbf{14.79} \\
        \midrule
        AM & RT-10 & 12.51 & 14.59 & 51.44 & \textbf{5.37} \\
        SC & RT-10 & 78.93 & 83.31 & 9.2 & \textbf{8.01} \\
        SCR & RT-10 & 18.74 & 24.64 & 9.96 & \textbf{8.5} \\
        \midrule
        AM & RT-3 & 28.82 & 33.03 & 80.06 & \textbf{12.75} \\
        SC & RT-3 & 82.12 & 86.03 & 16.93 & \textbf{12.23} \\
        SCR & RT-3 & 29.09 & 35.47 & 18.45 & \textbf{12.41} \\
        \bottomrule
    \end{tabular}
\end{table}

As for DD (see Table~\ref{tab:resultsAll}), both Tent Adaptation and Norm adaptation showed a low prediction error rate when processing AM and SCR for test-time adaptation; however, they produced much higher prediction error rates, with $77.32\%$ and $82.74\%$ respectively, when processing SC. On the contrary, TTT had a low error rate for SC and SCR, but obtained a high error rate (51.46\%) for AM. \textit{CoNMix}, on the other hand, \textit{is the most stable TTA method having obtained relatively low adaptation error rates of 6.27\%, 8.49\%, and 9.42\% for AM, SC, and SCR, respectively}. Tent Adaptation, Norm Adaptation, TTT, and CoNMix had a similar performance for the other two types of background noise (see Table~\ref{tab:resultsAll} for EB and RT noise conditions). 

\subsection{SNR at 3.0 dB Level}
\noindent Table~\ref{tab:resultsAll} also demonstrates the performance of all four TTA models with a 3.0 dB SNR level for three different types of background noise, including DD, EB, and RT, respectively. Due to the noisier SNR level (3.0 dB), the TTA prediction error rates increased across the board for all the models over all the datasets; see the visual comparison of Table~\ref{tab:resultsAll}. The exact error rate increase is summarised in Table~\ref{tab:error_rate_increase}.

All four TTA methods except CoNMix have suffered significantly from the increase in background noise level, with their adaptation performances being the worst on AM (i.e., a minimum of $26.72\%$ increase in error rate). CoNMix, on the other hand, demonstrated a consistent performance where the increase in error rate among the three datasets was similar and below $7\%$. 
\begin{table}[h]
    \centering
    \caption{Error rate increased under DD background noise between 10.0 dB and 3.0 dB SNR levels on AM, SC, and SCR.}
    \begin{tabular}{ccccl}
        \toprule
        Dataset & Tent & Norm & TTT & CoNMix \\
        \midrule
        AM & 29.14 & 26.72 & 31.96 & {\bf{6.58}} \\
        SC & 5.43 & {\bf{3.13}} & 10.76 & 5.98 \\
        SCR & 15.25 & 16.69 & 10.52 & {\bf{6.87}} \\
        \bottomrule
    \end{tabular}
    \label{tab:error_rate_increase}
\end{table}

A similar pattern of increased error rates by the four TTA methods for their adaptation results at the 3 dB SNR level compared to the 10 dB SNR level was also observed for the other two background noise types (See Table~\ref{tab:resultsAll}). This indicates that \textbf{CoNMix is the most reliable TTA method among the four approaches we compared in this research when faced with an increasing SNR}.

\section{Evaluation}\label{sec:evaluation}
\subsection{Analysis of TTA Adaptation Performance}\label{subsec:TTA_analysis}
\noindent To understand the adaptation performance of all four TTA methods on AM, SC, and SCR under different types of background noise with different noise levels, we compared each method's performance with that of their respective pre-trained model without adaptation, as shown in Table~\ref{tab:resultsAll_with_noAdaptation}.

Similar to the performance levels observed in Table~\ref{tab:resultsAll}, we can see that there is little difference between the model behaviours on different types of background noise (see Table~\ref{tab:resultsAll_with_noAdaptation}). However, the model behaviours on different datasets are in stark-contrast, as shown in Table~\ref{tab:resultsAll_with_noAdaptation}. Specifically, for AM, all four methods except TTT have significantly reduced the prediction error rate after adaptation compared to the high error rates produced by their respective pre-trained model without adaptation (see Table~\ref{tab:resultsAll_with_noAdaptation}). Note that the error rates without adaptation among all four methods are high but fairly comparable to each other, and \textbf{CoNMix is by far the most effective adaptation method by reducing the error rate to around 12-15\% and 5-7\% for 3 dB and 10 dB SNR levels, respectively}. TTT, on the other hand, did not demonstrate any significant advantage of using adaptation over the pre-trained model without adaptation.

For SC, all four methods except CoNMix have shown a negative adaptation effect, i.e., the prediction error rate after adaptation is higher than that without any adaptation (see Table~\ref{tab:resultsAll_with_noAdaptation}). Both versions of TENT, i.e., Tent Adaptation and Norm Adaptation had a much worse negative adaptation effect compared to TTT. Note that TTT had much lower error rates without any adaptation compared to TENT. Even though TTT exhibited a negative adaptation effect, its error rates were still comparable to the best performing method, CoNMix.

Clearly, SC did not benefit much from TTA or even suffered from negative adaptation in the case of TENT and TTT, which is very different from the results obtained on AM. To further investigate the issue, we conducted TTA with all four methods on SCR, a randomly selected subset of SC to reduce the number of classes in SC (see details in the Datasets sub-section in the Methodology section~\ref{sec:methodology}), and the results are shown in Table~\ref{tab:resultsAll_with_noAdaptation}. Interestingly, negative adaptation did not occur with TENT but did manifest in the case of TTT. The adaptation performance of CoNMix on SCR is similar to its performance on SC.

Since there is likely a change of data distribution between SC and SCR due to subsampling of classes, the change of TTA model behaviours are likely related to the distribution change. Furthermore, as TENT (including both Tent Adaptation and Norm Adaptation) only updates BatchNorm statistics, its opposite adaptation behaviours on AM and SC motivated us to investigate the data distribution of the two datasets, respectively, in the following section.

\begin{table*}
    \centering
    \caption{TTA results in terms of prediction error rates for all models (including no adaptation) on AM, SC, and SCR, under background noise conditions EB, DD, and TR, respectively, at both 3 dB and 10 dB SNR levels. Specifically, Tent, Norm, TTT, and CoNMix represent non-adapted error rates, while Tent-Ad, Norm-Ad, TTT-Ad, and CoNMix-Ad represent adapted error rates. The up ($\uparrow$) and down ($\downarrow$) arrow beside the adapted error rates presents the error rate increase or decrease compared with the non-adapted error rates, respectively.}
    \label{tab:resultsAll_with_noAdaptation}
    \begin{tabular}{cccccccccl}
        \toprule
        Dataset & Noise-SNR & Tent & Tent-Ad & Norm & Norm-Ad & TTT & TTT-Ad & CoNMix & CoNMix-Ad \\
        \midrule
        AM & EB-3 & 80.64 & 33.63$\downarrow$ & 80.64 & 35.92$\downarrow$ & 89.20 & 89.81$\uparrow$ & 70.09 & \textbf{15.01}$\downarrow$ \\
        AM & EB-10 & 69.40 & 12.44$\downarrow$ & 69.40 & 15.63$\downarrow$ & 55.36 & 51.46$\downarrow$ & 50.3 & \textbf{5.31}$\downarrow$ \\
        AM & DD-3 & 85.50 & 86.37$\uparrow$ & 85.50 & 86.06$\uparrow$ & 87.10 & 85.81$\downarrow$ & 65.67 & \textbf{12.85}$\downarrow$ \\
        AM & DD-10 & 67.86 & 17.23$\downarrow$ & 67.86 & 19.34$\downarrow$ & 66.47 & 53.85$\downarrow$ & 41.43 & \textbf{6.37}$\downarrow$ \\
        AM & RT-3 & 71.23 & 28.82$\downarrow$ & 71.23 & 33.03$\downarrow$ & 85.19 & 80.06$\downarrow$ & 71.90 & \textbf{12.75}$\downarrow$ \\
        AM & RT-10 & 67.34 & 12.51$\downarrow$ & 67.34 & 14.59$\downarrow$ & 53.98 & 51.44$\downarrow$ & 51.45 & \textbf{5.37}$\downarrow$ \\
        \midrule
        SC & EB-3 & 52.00 & 83.48$\uparrow$ & 52.00 & 86.20$\uparrow$ & 15.67 & 21.64$\uparrow$ & 22.77 & \textbf{14.13}$\downarrow$ \\
        SC & EB-10 & 57.53 & 79.20$\uparrow$ & 57.53 & 83.88$\uparrow$ & 8.53 & 10.86$\uparrow$ & 12.19 & \textbf{8.32}$\downarrow$ \\
        SC & DD-3 & 51.63 & 82.75$\uparrow$ & 51.63 & 85.87$\uparrow$ & 17.82 & 19.93$\uparrow$ & 23.77 & \textbf{14.47}$\downarrow$ \\
        SC & DD-10 & 33.07 & 77.32$\uparrow$ & 33.02 & 82.74$\uparrow$ & 7.78 & 9.17$\uparrow$ & 11.51 & \textbf{8.49}$\downarrow$ \\
        SC & RT-3 & 43.47 & 82.12$\uparrow$ & 43.42 & 86.03$\uparrow$ & 13.82 & 16.93$\uparrow$ & 19.00 & \textbf{12.21}$\downarrow$ \\
        SC & RT-10 & 34.48 & 78.93$\uparrow$ & 34.48 & 83.31$\uparrow$ & 8.00 & 9.20$\uparrow$ & 11.89 & \textbf{8.03}$\downarrow$ \\
        \midrule
        SCR & EB-3 & 48.48 & 34.27$\downarrow$ & 48.48 & 41.43$\downarrow$ & 20.08 & 24.98$\uparrow$ & 23.29 & \textbf{14.79}$\downarrow$ \\
        SCR & EB-10 & 33.30 & 19.78$\downarrow$ & 33.30 & 26.41$\downarrow$ & 9.00 & 12.04$\uparrow$ & 12.32 & \textbf{9.02}$\downarrow$ \\
        SCR & DD-3 & 45.07 & 33.52$\downarrow$ & 45.07 & 39.66$\downarrow$ & 19.99 & 19.87$\downarrow$ & 26.47 & \textbf{16.29}$\downarrow$ \\
        SCR & DD-10 & 26.68 & 18.27$\downarrow$ & 26.68 & 22.97$\downarrow$ & 8.86 & 9.35$\uparrow$ & 12.77 & \textbf{9.42}$\downarrow$ \\
        SCR & RT-3 & 41.11 & 29.01$\downarrow$ & 41.11 & 35.47$\downarrow$ & 15.63 & 18.45$\uparrow$ & 18.46 & \textbf{12.41}$\downarrow$ \\
        SCR & RT-10 & 21.01 & 18.74$\downarrow$ & 31.01 & 24.64$\downarrow$ & 8.10 & 9.96$\uparrow$ & 11.07 & \textbf{8.50}$\downarrow$\\
        \bottomrule
    \end{tabular}
\end{table*}

\subsection{Further Probe into the Performance of TENT}
\begin{figure}[h]
    \centering
    \includegraphics[width=1.0\linewidth]{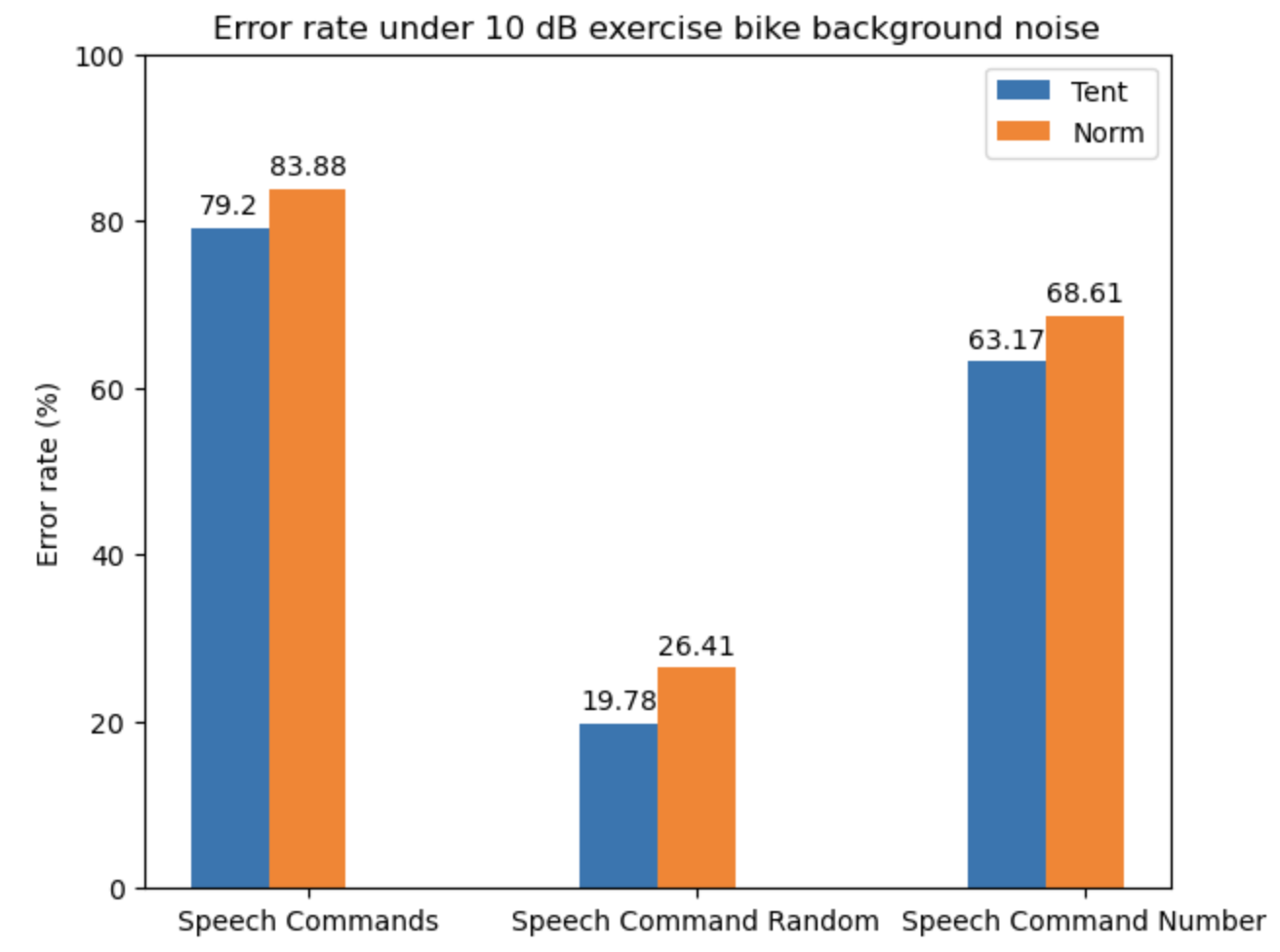}
    \caption{Demonstrates the Tent Adaptation and Norm Adaptation error rates for SC, SCR, and SCN, under 10 dB SNR level EB background noise.}
    \label{fig:10dB_exercise-bike_evaluation}
\end{figure}
\noindent As discussed in the previous section, TENT exhibited an improvement on SCR compared to its performance on AM and SC. We further investigated its performance on SC variants, including the original SC, SCR, and SCN (which contains the speech samples of the same 10 digits in English as found in AM). The results in terms of prediction error rates by both Tent Adaptation and Norm Adaptation are shown in Fig.\ref{fig:10dB_exercise-bike_evaluation} for the adaptation to EB noise. The performances of the two versions of TENT are similar to each other and both had the best performance on SCR, followed by SCN and SC.

As the authors for TENT~\cite{wang2021tent} designed domain shift with additive Gaussian noise in their experiments, we also further explore the application of TENT on audio datasets with Gaussian noise added to the test sets of AM, SC, SCR, and SCN. The Gaussian distribution shifts (Eq.\ref{eq:guas_shft}) were applied as follows:
\begin{equation}\label{eq:guas_shft}
    x = \mathbf{x} + \lambda \times \text{guassian\_noise}(\mathbf{x})
\end{equation}
where $\mathbf{x} \in \mathcal{X_T}$; $x \in \mathcal{X_T}$; $\text{guassian\_noise}(\mathbf{x})$ is the Gaussian noise of the same data shape as $\mathbf{x}$. The Gaussian noise is randomly generated with $\mu = 0$ and $\sigma = 1$; $\lambda \in [0, 1]$ controls the Gaussian noise level. 

\begin{figure}[h]
    \centering
    \includegraphics[width=1.0\linewidth]{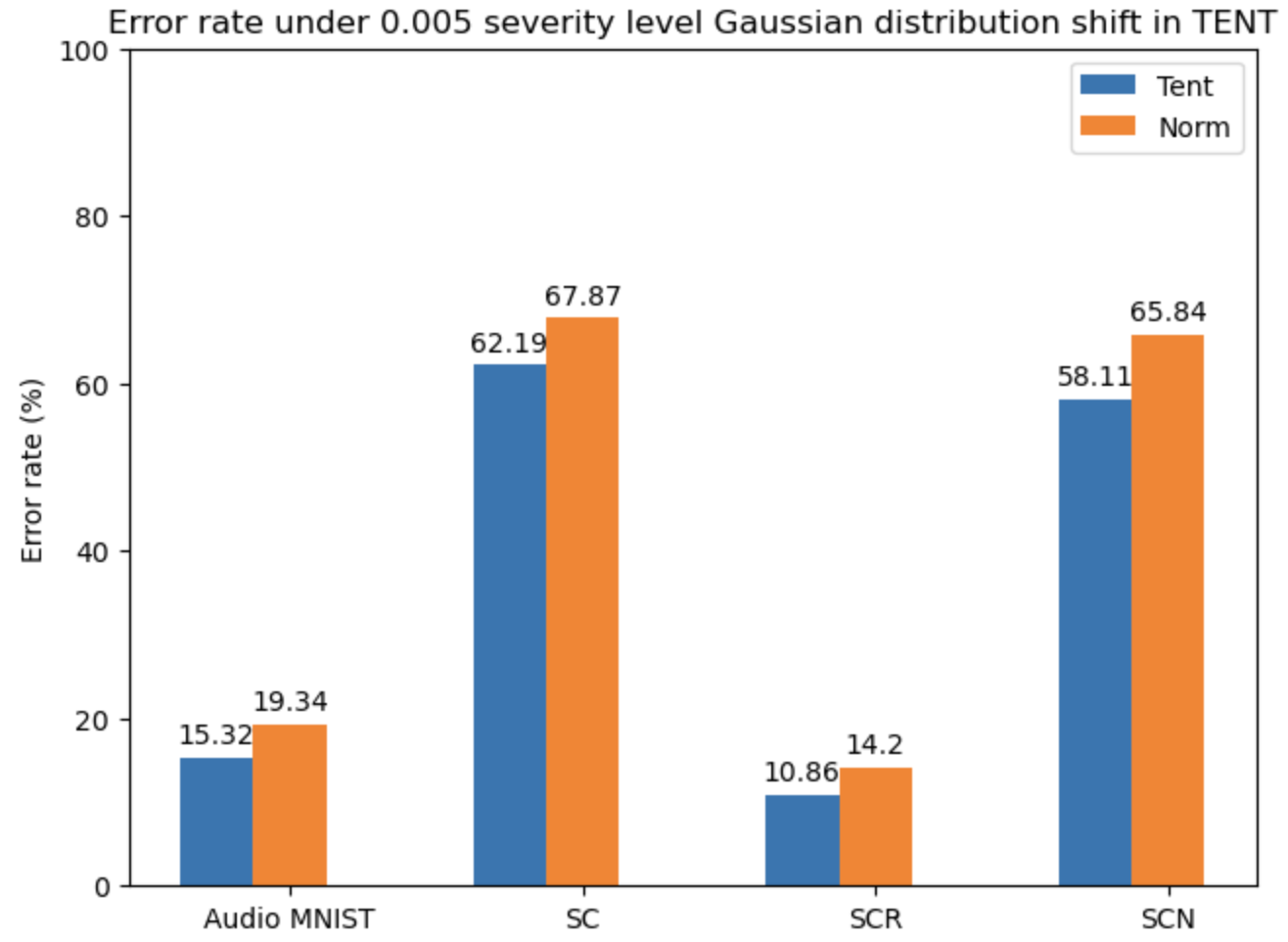}
    \caption{Includes the Tent Adaptation and Norm Adaptation error rates across four datasets, namely, AM, SC, SCR, and SCN, under 0.005 severity Gaussian distribution shift.}
    \label{fig:005_Gaussian_shift_in_TENT}
\end{figure}

The results of TENT adaptation are shown in Fig.\ref{fig:005_Gaussian_shift_in_TENT}. Both Tent Adaptation and Norm Adaptation had a similar performance to the scenario of adding background noise 'bike exercise' and again displayed much worse prediction error rates on the SC (62.19\% and 67.87\%) and SCN (58.11\% and 65.84\%) compared to that on AM (15.32\% and 19.34\%) and SCR (10.86\% and 14.20\%). This indicates that \textbf{the performance of TENT is not affected significantly by the types of background noise or Gaussian noise, but instead is strongly associated with the domain shift patterns between the training and test samples in a dataset}.

\subsection{TENT vs. TTT vs. CoNMix}\label{subsec:model_comparing}
\begin{figure}[h]
    \centering
    \includegraphics[width=1.0\linewidth]{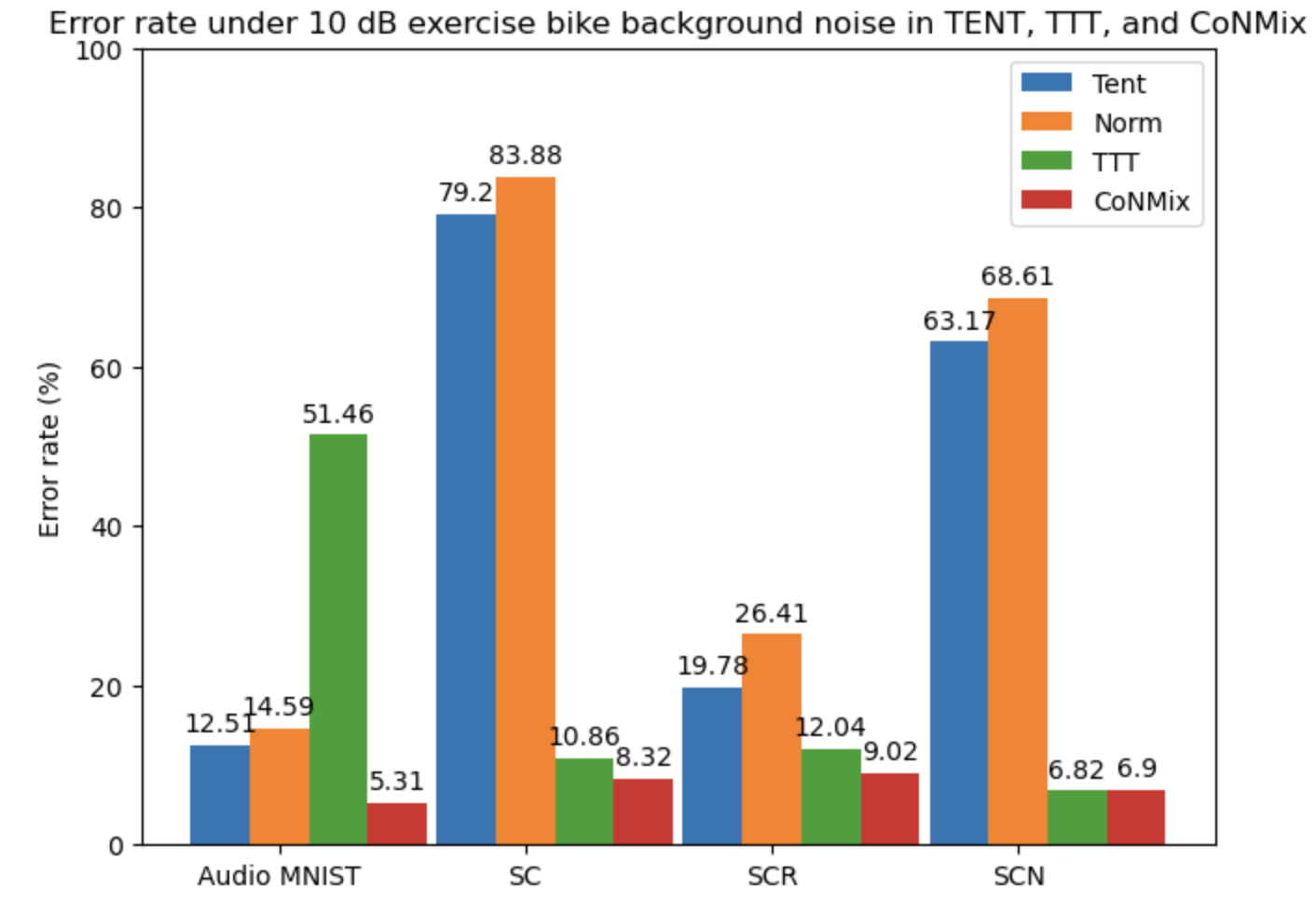}
    \caption{Shows error rates for the 10 dB SNR level EB background noise condition for TENT, TTT, and CoNMix across AM, SC, SCR, and SCN}
    \label{fig:10dB_exercise-bike_full_evaluation}
\end{figure}
\noindent We further investigate the difference between the adaptation performance levels of TENT, TTT, and CoNMix in this section. Firstly, for easy comparison, we summarise the adaptation results of all models on AM, SC, and SCR, as previously discussed, together with the results on SCN in Fig.\ref{fig:10dB_exercise-bike_full_evaluation}. It is clear that TENT performed poorly on SC (error rate $>79\%$) and SCN (error rate $>63\%$), whereas TTT performed badly on AM (error rate 51.46\%). CoNMix had a consistent performance across all four datasets with much lower error rates varying from 5.31\% to 9.02\%.

The superior performance by CoNMix can be attributed to its following characteristics:
\begin{enumerate}
    \item\label{item:CoNMix_multipleEpochs} CoNMix inherits the adapted weights and adapts for many epochs with the entire test set, whereas TENT and TTT process the test set in one epoch.
    \item\label{item:CoNMix_ViT} CoNMix adopts the Vision Transformer architecture which is more representationally expressive than the CNN model architectures adopted by TENT and TTT.
\end{enumerate}

\begin{figure}[h]
    \centering
    \includegraphics[width=1.0\linewidth]{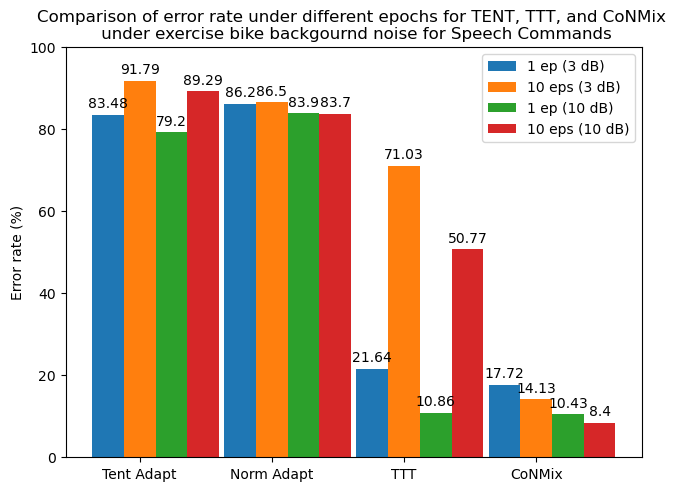}
    \caption{Demonstrates the test-time adaptation error rate at one epoch for 3 dB SNR level (1 ep 3 dB), ten epochs for 3 dB SNR level (10 eps 3 dB), one epoch for 10 dB SNR level (1 ep 10 dB), and ten epochs for 10 dB SNR level (10 eps 10 dB) under exercise bike noise.}
    \label{fig:multi-epoch_exercise-bike_Speech-Commands}
\end{figure}

For point \ref{item:CoNMix_multipleEpochs}, one may think intuitively that tuning TENT and TTT at test time with the full test set for multiple epochs may benefit their adaptation performance. To answer this question, we conducted the experiment to train TENT, TTT, and CoNMix at test time for the same number of epochs and recorded the prediction error rates using the adapted models at 1 epoch and 10 epochs, respectively. The results for all the models on SC with EB are shown in Fig.\ref{fig:multi-epoch_exercise-bike_Speech-Commands}. Interestingly, we observe that \textbf{TENT (both Tent and Norm) and TTT did not benefit from training with multiple epochs at test time}. This indicates that the issue lies with the fundamental design of TENT and TTT, where only one sample or a mini-batch of samples is used for updating model parameters, which is then used for predicting the same sample or mini-batch. As discussed previously, this updating process is stochastic and hence unstable for reliable immediate predictions. However, OTTA or TTBA are appealing for real-time adaptation and prediction at test time. So it is challenging but beneficial to research how to improve the adaptation performance with OTTA or TTBA settings.

\begin{figure}[h]
    \centering
    \includegraphics[width=1.0\linewidth]{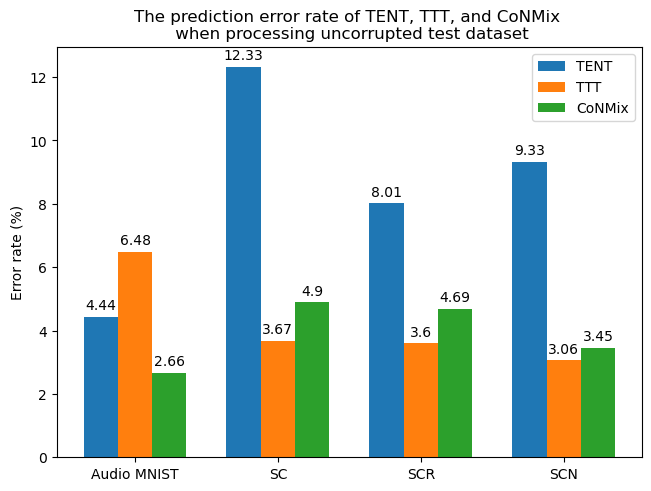}
    \caption{Depicts error rates when processing the uncorrupted test sets without the adaptation across AM, SC, SCR, and SCN. The results show that CoNMix is the most stable model since the error rate exhibits the lowest fluctuation.}
    \label{fig:no-corrupted_test_dataset_evaluation}
\end{figure}

For point \ref{item:CoNMix_ViT}, we look closely into the architectures of TENT, TTT, and CoNMix:
\begin{itemize}
    \item {\bf{TENT}} adopts the standard ResNet50 architecture which consists of BatchNorm, ReLU, and 2D Convolutional layers. Note that Tent Adaptation and Norm Adaptation have the same model architecture.
    \item {\bf{TTT}} leverages Transfer Learning from a ResNet architecture but only includes 26 layers of BatchNorm, ReLU, and 2D Convolutional layers. Compared to TENT, the main difference is that TTT has two classification heads which share the same ResNet backbone. 
    \item {\bf{CoNMix}} is a much larger model compared to TENT and TTT, consisting of an embedding network and a transformer network. The embedding network is ResNet50 with 50 layers, however, BatchNorm is replaced with GroupNorm~\cite{wu2018groupnormalization}. The transformer network includes 12 multi-head attention blocks (see section~\ref{sec:methodology} for details).
\end{itemize}
\textbf{CoNMix has an obvious advantage in terms of a strong feature representation compared to TENT and TTT due to the superior performance of transformers in a wide range of applications}. To see the advantage of CoNMix, we conducted an experiment where pre-trained models of TENT, TTT, and CoNMix were applied to uncorrupted test samples without any adaptation across all four datasets: AM, SC, SCR, and SCN. The results are demonstrated in Fig.\ref{fig:no-corrupted_test_dataset_evaluation}. We observe that \textit{CoNMix still had the most stable performance with comparatively low error rates (between 2.66\% to 4.90\%) across all datasets}. TTT had the highest error rate on AM (which displays a large semantic domain gap) due to its least adaptable training strategy, as discussed in the previous sections. However, TTT performed the best on SC variants with error rates varying from 3.06\% to 3.67\%, attributed to the fact that the SC does not present any apparent domain shift between training and uncorrupted test sets. TENT, on the other hand, had the highest error rates on variants of SC (between 8.01\% to 12.33\%) due to its adaptation strategy which updates BatchNorm statistics in an online manner on mini-batches.

\section{Conclusion}

\noindent We investigated three typical TTA methods which were originally designed for Computer Vision tasks, including TENT, TTT, and CoNMix, where during test time adaptation, TENT adapts to batch normalisation statistics, TTT updates a shared backbone through a pretext task head, and CoNMix updates the entire model parameters via a complex combination of TTA strategies. Both TENT and TTT adapt model parameters based on test set mini-batches and use the updated model's parameters for immediate predictions for the same mini-batches, whereas CoNMix updates model parameters using the entire test set before making predictions for the entire test set.

For Q1 in section~\ref{sec:introduction}, we have reused the TTA methods, such as TENT, TTT, and CoNMix that were designed for Computer Vision, and our upgraded CoNMix contributes the best performance (see section~\ref{sec:experiment}). However, the usage of TENT and TTT is unsuccessful for Audio Classification since TENT and TTT have a negative adaptation effect for SC (see subsection~\ref{subsec:TTA_analysis} in section~\ref{sec:evaluation}).

For Q2 in section~\ref{sec:introduction}, this research addresses TTA methods to process Audio Classification tasks for AM and SC. In section~\ref{sec:experiment}, the experimental results indicate that CoNMix has the highest performance when applying TTA methods at 3 dB and 10 dB SNR levels for AM and SC. As for our degradation in CoNMix, subsection~\ref{subapp:as_for_sc} in Appendix~\ref{app:ablation_study} indicates Pseudo-label loss (\ref{CoNMix_pl_ce_loss}) leads to a negative adaptation effect and NLL loss (\ref{CoNMix_pl_nll_loss}) solves the negative effect when processing TTA methods for SC. However, for the best solution, we omit the Pseudo-label loss (\ref{CoNMix_pl_ce_loss}) when processing TTA methods for AM (see subsection~\ref{subapp:as_for_am} in Appendix~\ref{app:ablation_study}). In section~\ref{sec:evaluation} subsection~\ref{subsec:TTA_analysis}, TENT and TTT are shown to be ineffective since they have a negative adaptation effect when applying TTA methods for SC (see Table~\ref{tab:resultsAll_with_noAdaptation} in section~\ref{sec:evaluation}). 


\section*{Acknowledgments}
\noindent The authors would like to thank Google Cloud Research for donating GPU credits for running the experiments reported in this paper.

\begin{small}
\bibliographystyle{plain}
\bibliography{reference}

\begin{thebibliography}{10}

\bibitem{Amiri2024PathologySpeechDetection}
Mahdi Amiri and Ina Kodrasi.
\newblock Test-time adaptation for automatic pathological speech detection in noisy environments.
\newblock In {\em 2024 32nd European Signal Processing Conference (EUSIPCO)}, pages 86--90, 2024.

\bibitem{azimi2022self}
Fatemeh Azimi, Sebastian Palacio, Federico Raue, J{\"o}rn Hees, Luca Bertinetto, and Andreas Dengel.
\newblock Self-supervised test-time adaptation on video data.
\newblock In {\em Proceedings of the IEEE/CVF Winter Conference on Applications of Computer Vision}, pages 3439--3448, 2022.

\bibitem{audiomnist2023}
Sören Becker, Johanna Vielhaben, Marcel Ackermann, Klaus-Robert Müller, Sebastian Lapuschkin, and Wojciech Samek.
\newblock Audiomnist: Exploring explainable artificial intelligence for audio analysis on a simple benchmark.
\newblock {\em Journal of the Franklin Institute}, 2023.

\bibitem{borlino2022self}
Francesco~Cappio Borlino, Salvatore Polizzotto, Barbara Caputo, and Tatiana Tommasi.
\newblock Self-supervision \& meta-learning for one-shot unsupervised cross-domain detection.
\newblock {\em Computer Vision and Image Understanding}, 223:103549, 2022.

\bibitem{chen2022source}
Peipeng Chen and Andy~J Ma.
\newblock Source-free temporal attentive domain adaptation for video action recognition.
\newblock In {\em Proceedings of the 2022 International Conference on Multimedia Retrieval}, pages 489--497, 2022.

\bibitem{dosovitskiy2020image}
Alexey Dosovitskiy.
\newblock An image is worth 16x16 words: Transformers for image recognition at scale.
\newblock {\em arXiv preprint arXiv:2010.11929}, 2020.

\bibitem{dubey2021adaptive}
Abhimanyu Dubey, Vignesh Ramanathan, Alex Pentland, and Dhruv Mahajan.
\newblock Adaptive methods for real-world domain generalization.
\newblock In {\em Proceedings of the IEEE/CVF Conference on Computer Vision and Pattern Recognition}, pages 14340--14349, 2021.

\bibitem{farahani2021brief}
Abolfazl Farahani, Sahar Voghoei, Khaled Rasheed, and Hamid~R Arabnia.
\newblock A brief review of domain adaptation.
\newblock {\em Advances in data science and information engineering: proceedings from ICDATA 2020 and IKE 2020}, pages 877--894, 2021.

\bibitem{finn2017model}
Chelsea Finn, Pieter Abbeel, and Sergey Levine.
\newblock Model-agnostic meta-learning for fast adaptation of deep networks.
\newblock In {\em International conference on machine learning}, pages 1126--1135. PMLR, 2017.

\bibitem{goodfellow2016deep}
Ian Goodfellow, Yoshua Bengio, and Aaron Courville.
\newblock {\em Deep learning}.
\newblock MIT press, 2016.

\bibitem{he2016deep}
Kaiming He, Xiangyu Zhang, Shaoqing Ren, and Jian Sun.
\newblock Deep residual learning for image recognition.
\newblock In {\em Proceedings of the IEEE conference on computer vision and pattern recognition}, pages 770--778, 2016.

\bibitem{hospedales2021meta}
Timothy Hospedales, Antreas Antoniou, Paul Micaelli, and Amos Storkey.
\newblock Meta-learning in neural networks: A survey.
\newblock {\em IEEE transactions on pattern analysis and machine intelligence}, 44(9):5149--5169, 2021.

\bibitem{hou2021visualizing}
Yunzhong Hou and Liang Zheng.
\newblock Visualizing adapted knowledge in domain transfer.
\newblock In {\em Proceedings of the IEEE/CVF conference on computer vision and pattern recognition}, pages 13824--13833, 2021.

\bibitem{hwang2020mel}
Yeongtae Hwang, Hyemin Cho, Hongsun Yang, Dong-Ok Won, Insoo Oh, and Seong-Whan Lee.
\newblock Mel-spectrogram augmentation for sequence to sequence voice conversion.
\newblock {\em arXiv preprint arXiv:2001.01401}, 2020.

\bibitem{kim2023sgem}
Changhun Kim, Joonhyung Park, Hajin Shim, and Eunho Yang.
\newblock Sgem: Test-time adaptation for automatic speech recognition via sequential-level generalized entropy minimization.
\newblock {\em arXiv preprint arXiv:2306.01981}, 2023.

\bibitem{kim2022variational}
Jangho Kim, Jun-Tae Lee, Simyung Chang, and Nojun Kwak.
\newblock Variational on-the-fly personalization.
\newblock In {\em International Conference on Machine Learning}, pages 11134--11147. PMLR, 2022.

\bibitem{kim2021test}
Sunwoo Kim and Minje Kim.
\newblock Test-time adaptation toward personalized speech enhancement: Zero-shot learning with knowledge distillation.
\newblock In {\em 2021 IEEE Workshop on Applications of Signal Processing to Audio and Acoustics (WASPAA)}, pages 176--180. IEEE, 2021.

\bibitem{klejch2019lattice}
Ondrej Klejch, Joachim Fainberg, Peter Bell, and Steve Renals.
\newblock Lattice-based unsupervised test-time adaptation of neural network acoustic models.
\newblock {\em arXiv preprint arXiv:1906.11521}, 2019.

\bibitem{koniusz2017domain}
Piotr Koniusz, Yusuf Tas, and Fatih Porikli.
\newblock Domain adaptation by mixture of alignments of second-or higher-order scatter tensors.
\newblock In {\em Proceedings of the IEEE conference on computer vision and pattern recognition}, pages 4478--4487, 2017.

\bibitem{kumar2023conmix}
Vikash Kumar, Rohit Lal, Himanshu Patil, and Anirban Chakraborty.
\newblock Conmix for source-free single and multi-target domain adaptation.
\newblock In {\em Proceedings of the IEEE/CVF Winter Conference on Applications of Computer Vision}, pages 4178--4188, 2023.

\bibitem{li2020model}
Rui Li, Qianfen Jiao, Wenming Cao, Hau-San Wong, and Si~Wu.
\newblock Model adaptation: Unsupervised domain adaptation without source data.
\newblock In {\em Proceedings of the IEEE/CVF conference on computer vision and pattern recognition}, pages 9641--9650, 2020.

\bibitem{li2021imbalanced}
Xinhao Li, Jingjing Li, Lei Zhu, Guoqing Wang, and Zi~Huang.
\newblock Imbalanced source-free domain adaptation.
\newblock In {\em Proceedings of the 29th ACM international conference on multimedia}, pages 3330--3339, 2021.

\bibitem{liang2024comprehensive}
Jian Liang, Ran He, and Tieniu Tan.
\newblock A comprehensive survey on test-time adaptation under distribution shifts.
\newblock {\em International Journal of Computer Vision}, pages 1--34, 2024.

\bibitem{lin2024continual}
Guan-Ting Lin, Wei-Ping Huang, and Hung-yi Lee.
\newblock Continual test-time adaptation for end-to-end speech recognition on noisy speech.
\newblock {\em arXiv preprint arXiv:2406.11064}, 2024.

\bibitem{lin2022listen}
Guan-Ting Lin, Shang-Wen Li, and Hung-yi Lee.
\newblock Listen, adapt, better wer: Source-free single-utterance test-time adaptation for automatic speech recognition.
\newblock {\em arXiv preprint arXiv:2203.14222}, 2022.

\bibitem{motiian2017unified}
Saeid Motiian, Marco Piccirilli, Donald~A Adjeroh, and Gianfranco Doretto.
\newblock Unified deep supervised domain adaptation and generalization.
\newblock In {\em Proceedings of the IEEE international conference on computer vision}, pages 5715--5725, 2017.

\bibitem{nelakurthi2018source}
Arun~Reddy Nelakurthi, Ross Maciejewski, and Jingrui He.
\newblock Source free domain adaptation using an off-the-shelf classifier.
\newblock In {\em 2018 IEEE International conference on big data (Big Data)}, pages 140--145. IEEE, 2018.

\bibitem{park2020fast}
Seobin Park, Jinsu Yoo, Donghyeon Cho, Jiwon Kim, and Tae~Hyun Kim.
\newblock Fast adaptation to super-resolution networks via meta-learning.
\newblock In {\em Computer Vision--ECCV 2020: 16th European Conference, Glasgow, UK, August 23--28, 2020, Proceedings, Part XXVII 16}, pages 754--769. Springer, 2020.

\bibitem{saito2019semi}
Kuniaki Saito, Donghyun Kim, Stan Sclaroff, Trevor Darrell, and Kate Saenko.
\newblock Semi-supervised domain adaptation via minimax entropy.
\newblock In {\em Proceedings of the IEEE/CVF international conference on computer vision}, pages 8050--8058, 2019.

\bibitem{shu2022test}
Manli Shu, Weili Nie, De-An Huang, Zhiding Yu, Tom Goldstein, Anima Anandkumar, and Chaowei Xiao.
\newblock Test-time prompt tuning for zero-shot generalization in vision-language models.
\newblock {\em Advances in Neural Information Processing Systems}, 35:14274--14289, 2022.

\bibitem{sun2020test}
Yu~Sun, Xiaolong Wang, Zhuang Liu, John Miller, Alexei Efros, and Moritz Hardt.
\newblock Test-time training with self-supervision for generalization under distribution shifts.
\newblock In {\em International conference on machine learning}, pages 9229--9248. PMLR, 2020.

\bibitem{tan2018survey}
Chuanqi Tan, Fuchun Sun, Tao Kong, Wenchang Zhang, Chao Yang, and Chunfang Liu.
\newblock A survey on deep transfer learning.
\newblock In {\em Artificial Neural Networks and Machine Learning--ICANN 2018: 27th International Conference on Artificial Neural Networks, Rhodes, Greece, October 4-7, 2018, Proceedings, Part III 27}, pages 270--279. Springer, 2018.

\bibitem{ustubioglu2023mel}
Arda Ustubioglu, Beste Ustubioglu, and Guzin Ulutas.
\newblock Mel spectrogram-based audio forgery detection using cnn.
\newblock {\em Signal, Image and Video Processing}, 17(5):2211--2219, 2023.

\bibitem{venkateswara2017deep}
Hemanth Venkateswara, Jose Eusebio, Shayok Chakraborty, and Sethuraman Panchanathan.
\newblock Deep hashing network for unsupervised domain adaptation.
\newblock In {\em Proceedings of the IEEE Conference on Computer Vision and Pattern Recognition}, pages 5018--5027, 2017.

\bibitem{wang2021tent}
Dequan Wang, Evan Shelhamer, Shaoteng Liu, Bruno Olshausen, and Trevor Darrell.
\newblock Tent: Fully test-time adaptation by entropy minimization.
\newblock In {\em International Conference on Learning Representations}, 2021.

\bibitem{wang2023test}
Renhao Wang, Yu~Sun, Yossi Gandelsman, Xinlei Chen, Alexei~A Efros, and Xiaolong Wang.
\newblock Test-time training on video streams.
\newblock {\em arXiv preprint arXiv:2307.05014}, 2023.

\bibitem{wang2021efficient}
Xinyi Wang, Yulia Tsvetkov, Sebastian Ruder, and Graham Neubig.
\newblock Efficient test time adapter ensembling for low-resource language varieties.
\newblock {\em arXiv preprint arXiv:2109.04877}, 2021.

\bibitem{warden2018speech}
Pete Warden.
\newblock Speech commands: A dataset for limited-vocabulary speech recognition.
\newblock {\em arXiv preprint arXiv:1804.03209}, 2018.

\bibitem{wu2018groupnormalization}
Yuxin Wu and Kaiming He.
\newblock Group normalization, 2018.

\bibitem{yao2015semi}
Ting Yao, Yingwei Pan, Chong-Wah Ngo, Houqiang Li, and Tao Mei.
\newblock Semi-supervised domain adaptation with subspace learning for visual recognition.
\newblock In {\em Proceedings of the IEEE conference on Computer Vision and Pattern Recognition}, pages 2142--2150, 2015.

\bibitem{ye2022alleviating}
Yalan Ye, Ziqi Liu, Yangwuyong Zhang, Jingjing Li, and Hengtao Shen.
\newblock Alleviating style sensitivity then adapting: Source-free domain adaptation for medical image segmentation.
\newblock In {\em Proceedings of the 30th ACM International Conference on Multimedia}, pages 1935--1944, 2022.

\bibitem{zeyang2022unsupervised}
Wang Zeyang.
\newblock An unsupervised domain adaptation method for compressed video quality enhancement.
\newblock In {\em 2022 19th International Computer Conference on Wavelet Active Media Technology and Information Processing (ICCWAMTIP)}, pages 1--5. IEEE, 2022.

\bibitem{zhang2021source}
Dan Zhang, Mao Ye, Lin Xiong, Shuaifeng Li, and Xue Li.
\newblock Source-style transferred mean teacher for source-data free object detection.
\newblock In {\em Proceedings of the 3rd ACM International Conference on Multimedia in Asia}, pages 1--8, 2021.

\bibitem{zhang2022transfer}
Lei Zhang and Xinbo Gao.
\newblock Transfer adaptation learning: A decade survey.
\newblock {\em IEEE Transactions on Neural Networks and Learning Systems}, 2022.

\end{thebibliography}
\end{small}

\appendix
\section{Ablation Study for CoNMix}\label{app:ablation_study}
\subsection{Ablation Analysis for SC}\label{subapp:as_for_sc}
\begin{figure*}[!t]
    \centering
    \subfloat[\label{fig:PA-10dB-RT-SC}]{\includegraphics[width=0.33\linewidth]{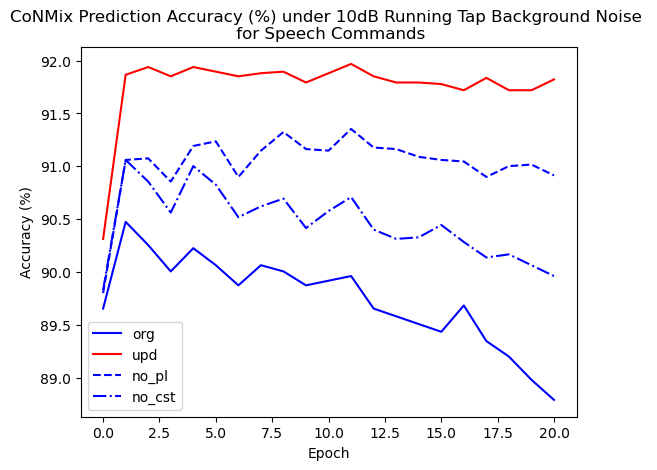}}
    \hfill
    \subfloat[\label{fig:PA-10dB-EB-SC}]{\includegraphics[width=0.33\linewidth]{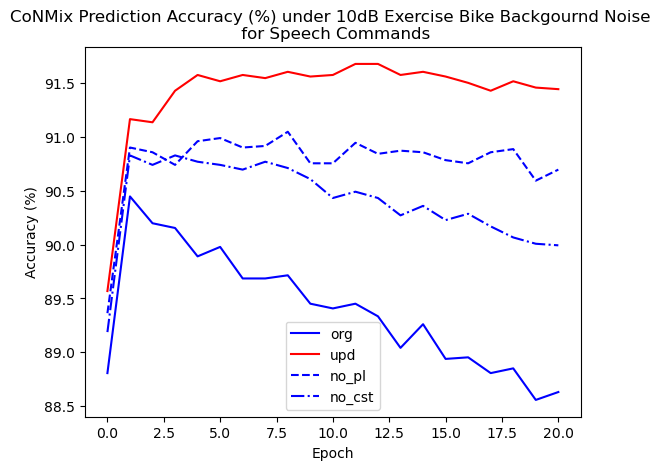}}
    \hfill
    \subfloat[\label{fig:PA-10dB-DD-SC}]{\includegraphics[width=0.33\linewidth]{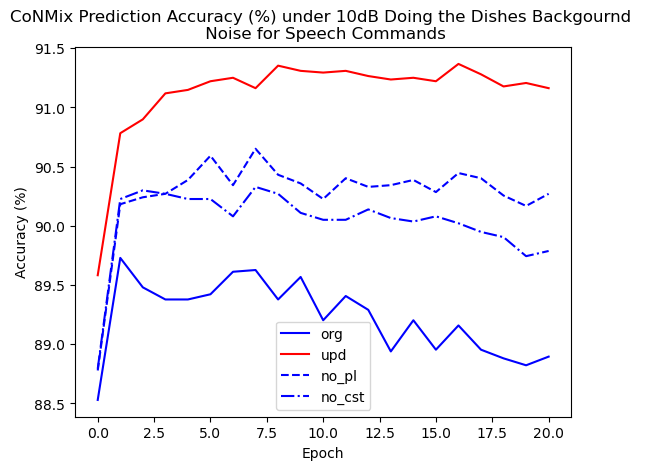}}
    \hfill
    \subfloat[\label{fig:PA-10dB-no_nm-SC}]{\includegraphics[width=0.33\linewidth]{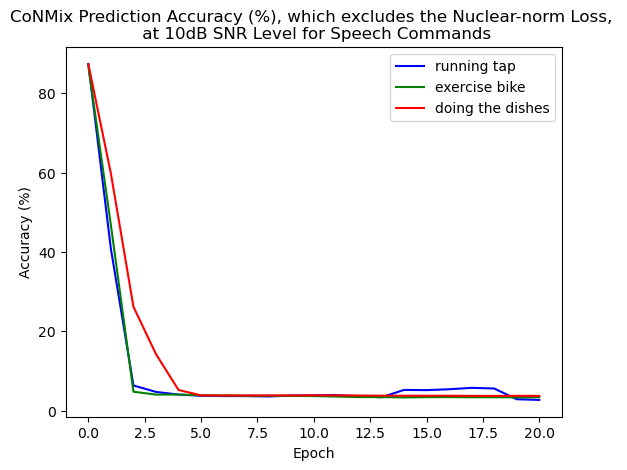}}
    \hfill
    \subfloat[\label{fig:PL-10dB-org-SC}]{\includegraphics[width=0.33\linewidth]{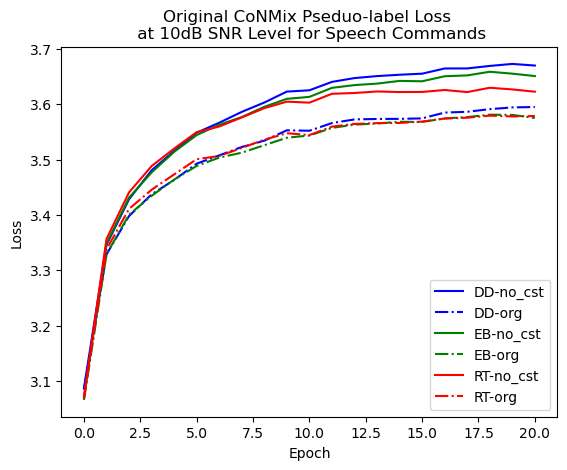}}
    \hfill
    \subfloat[\label{fig:PL-10dB-SC}]{\includegraphics[width=0.33\linewidth]{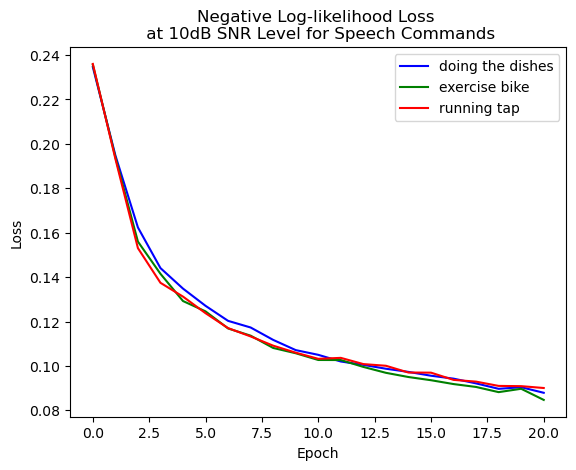}}
    \hfill
    \subfloat[\label{fig:PA-3dB-DD-SC}]{\includegraphics[width=0.33\linewidth]{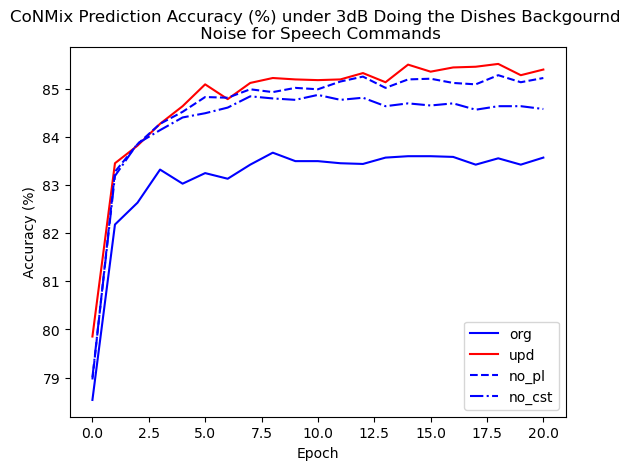}}
    \hfill
    \subfloat[\label{fig:PA-3dB-EB-SC}]{\includegraphics[width=0.33\linewidth]{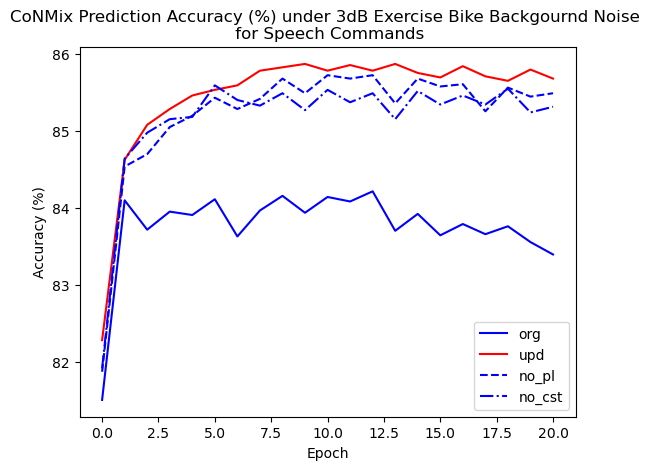}}
    \hfill
    \subfloat[\label{fig:PA-3dB-RT-SC}]{\includegraphics[width=0.33\linewidth]{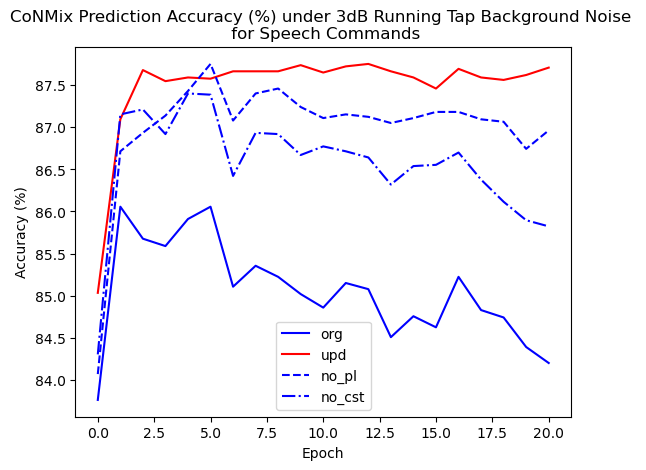}}
    \hfill
    \subfloat[\label{fig:PA-3dB-no_nm-SC.}]{\includegraphics[width=0.33\linewidth]{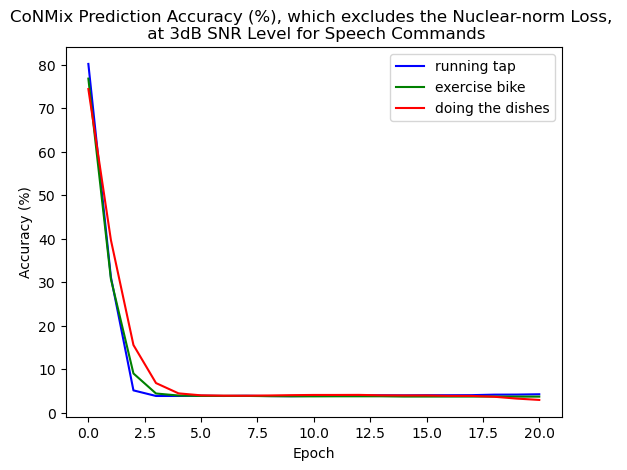}}
    \hfill
    \subfloat[\label{fig:PL-3dB-org-SC}]{\includegraphics[width=0.33\linewidth]{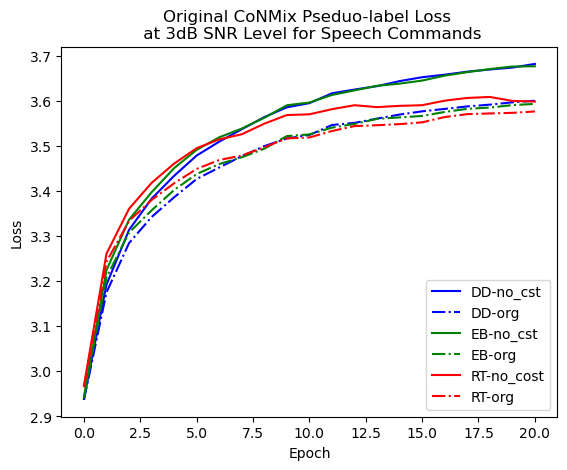}}
    \hfill
    \subfloat[\label{fig:PL-3dB-SC}]{\includegraphics[width=0.33\linewidth]{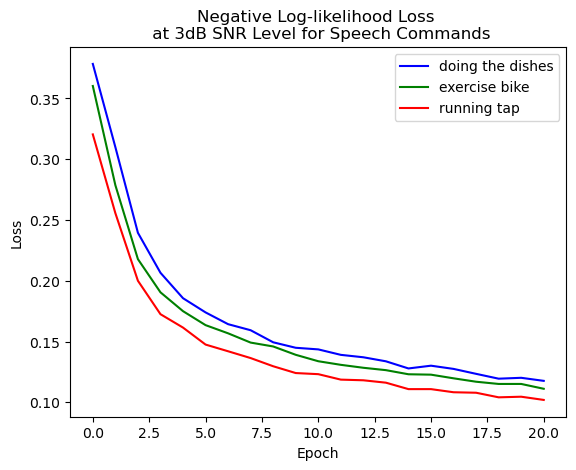}}
    \caption{The accuracy results at 10 dB ((a), (b), (c), and (d)) and 3 dB ((g), (h), (i), and (j)) SNR level for SC. To illustrate, upd (replace Pseudio-label loss (\ref{CoNMix_pl_ce_loss}) with NLL loss (\ref{CoNMix_pl_nll_loss})) (see \ref{sec:methodology})), org (original CoNMix Solution), no\_cst (exclude consistency loss (\ref{eq:L_cons})), no\_pl (exclude pseudo-label loss (\ref{CoNMix_pl_ce_loss})), and no\_nm (exclude Nuclear-norm Maximization loss (\ref{Nuclear_norm_loss})). CoNMix Pseudo-label Loss under three 10 dB ((e) and (f)) and 3 dB ((k) and (l)) SNR level background noises, such as DD ('doing the dishes'), EB ('exercise bike'), and RT ('running tap'), for SC.}
    \label{fig:allResult-PA-10dB-SC}
\end{figure*}

\noindent As shown in Fig.\ref{fig:PA-10dB-RT-SC},~\ref{fig:PA-10dB-EB-SC},~\ref{fig:PA-10dB-DD-SC},~\ref{fig:PA-3dB-RT-SC},~\ref{fig:PA-3dB-EB-SC}, and~\ref{fig:PA-3dB-DD-SC}, the original CoNMix pseudo-label loss (Eq.\ref{CoNMix_pl_ce_loss}) acts as a negative adaptation effect. If the CoNMix model omits the original pseudo-label loss, then the negative adaptation effect does not occur. Meanwhile, the Nuclear Norm loss (Eq.\ref{Nuclear_norm_loss}) plays a most significant role in CoNMix loss functions, if the CoNMix model omits (Eq.\ref{Nuclear_norm_loss}), the prediction accuracy will reduce to less than  $10 \%$ (see Fig.\ref{fig:PA-10dB-no_nm-SC} and~\ref{fig:PA-3dB-no_nm-SC.}).

Another important aspect is the original CoNMix pseudo-label loss (Eq.\ref{CoNMix_pl_ce_loss}) has an increased tendency (see Fig.\ref{fig:PL-10dB-org-SC} and~\ref{fig:PL-3dB-org-SC}). After replacing (Eq.\ref{CoNMix_pl_ce_loss}) with NLL loss (Eq.\ref{CoNMix_pl_nll_loss}), the pseudo-label loss is convergent (see Fig.\ref{fig:PL-10dB-SC} and~\ref{fig:PL-3dB-SC}).

\subsection{Ablation Analysis for AM}\label{subapp:as_for_am}
\begin{figure*}[!t]
    \centering
    \subfloat[\label{fig:PA-10dB-DD-AM}]{\includegraphics[width=0.33\linewidth]{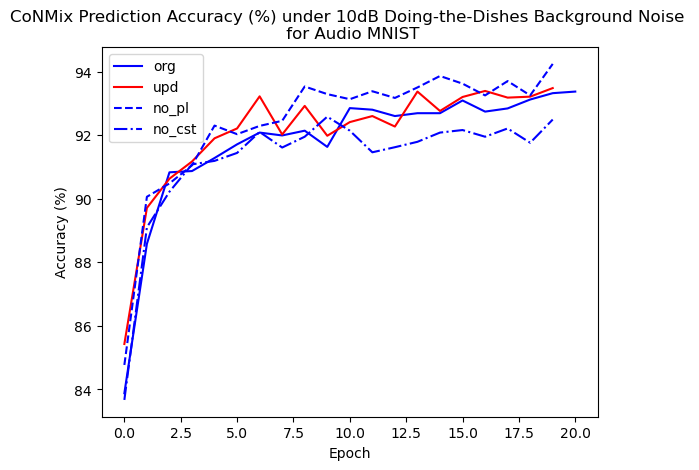}}
    \hfill
    \subfloat[\label{fig:PA-10dB-EB-AM}]{\includegraphics[width=0.33\linewidth]{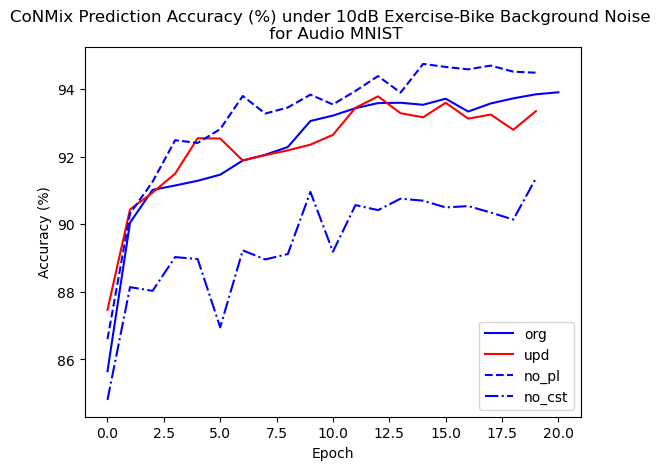}}
    \hfill
    \subfloat[\label{fig:PA-10dB-RT-AM}]{\includegraphics[width=0.33\linewidth]{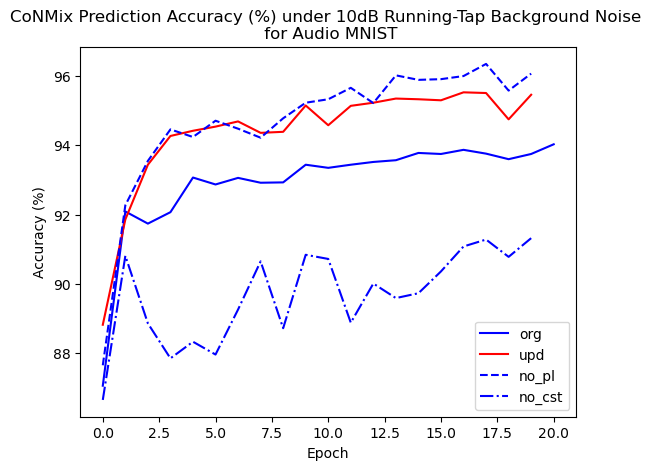}}
    \hfill
    \subfloat[\label{fig:PA-10dB-no_nm-AM}]{\includegraphics[width=0.33\linewidth]{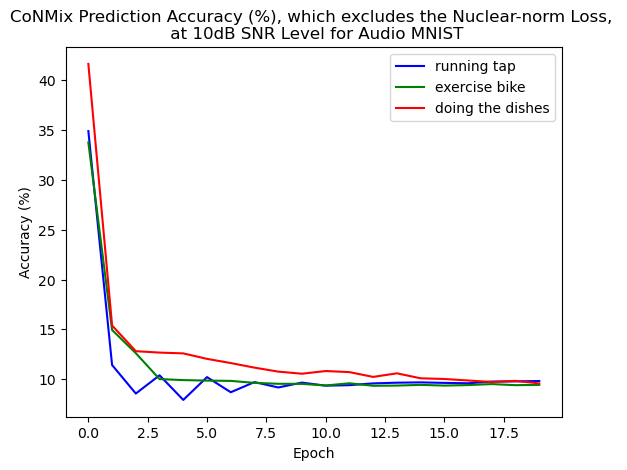}}
    \hfill
    \subfloat[\label{fig:PL-10dB-org-AM}]{\includegraphics[width=0.33\linewidth]{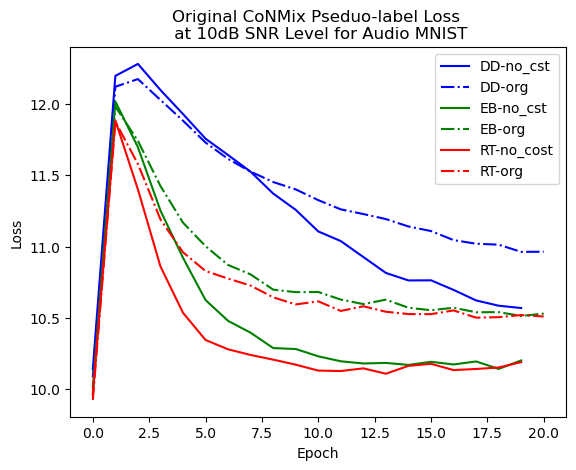}}
    \hfill
    \subfloat[\label{fig:PL-10dB-AM}]{\includegraphics[width=0.33\linewidth]{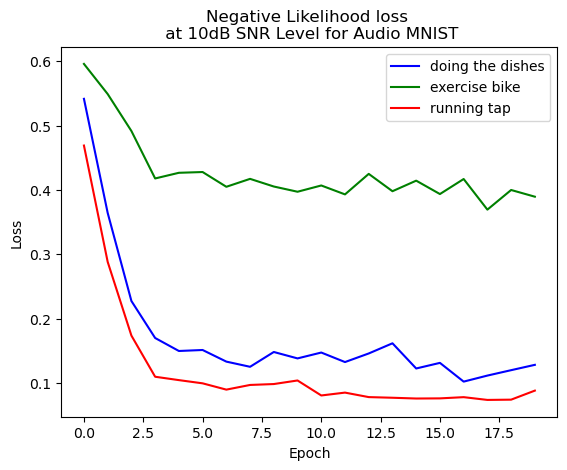}}
    \hfill
    \centering
    \subfloat[\label{fig:PA-3dB-DD-AM}]{\includegraphics[width=0.33\linewidth]{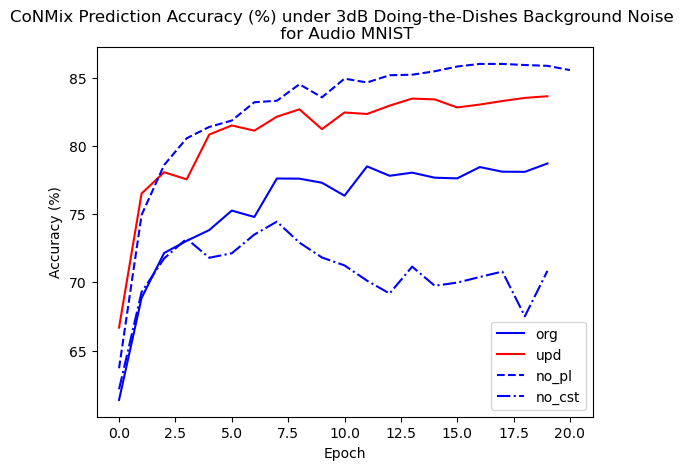}}
    \hfill
    \subfloat[\label{fig:PA-3dB-EB-AM}]{\includegraphics[width=0.33\linewidth]{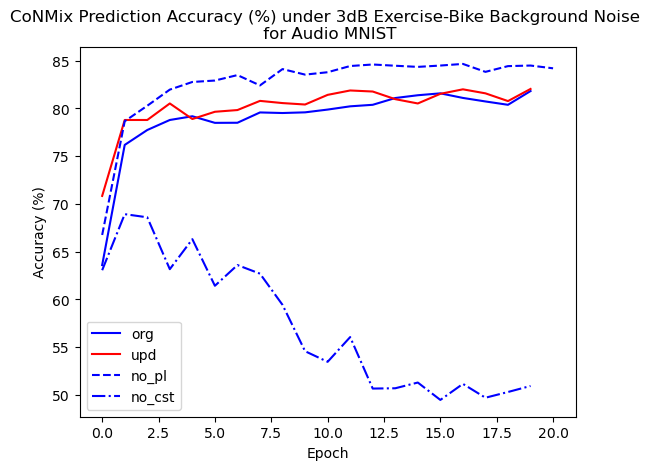}}
    \hfill
    \subfloat[\label{fig:PA-3dB-RT-AM}]{\includegraphics[width=0.33\linewidth]{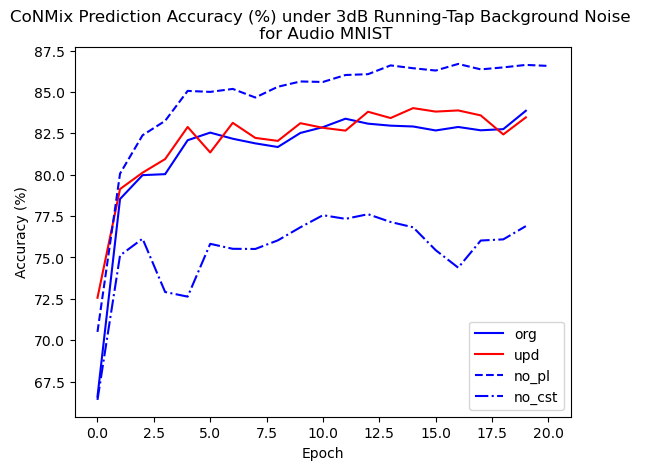}}
    \hfill
    \subfloat[\label{fig:PA-3dB-no_nm-AM}]{\includegraphics[width=0.33\linewidth]{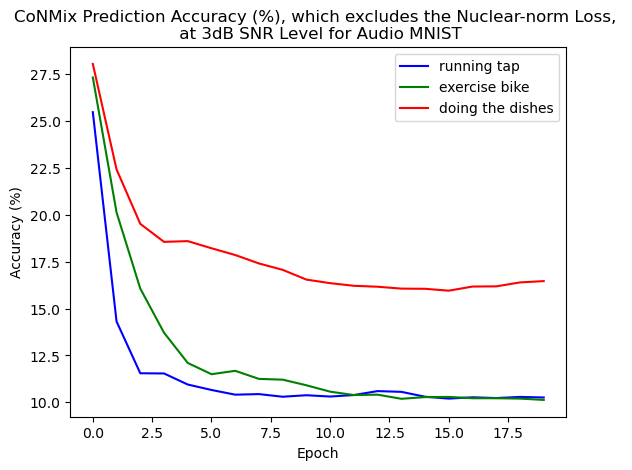}}
    \hfill
    \subfloat[\label{fig:PL-3dB-org-AM}]{\includegraphics[width=0.33\linewidth]{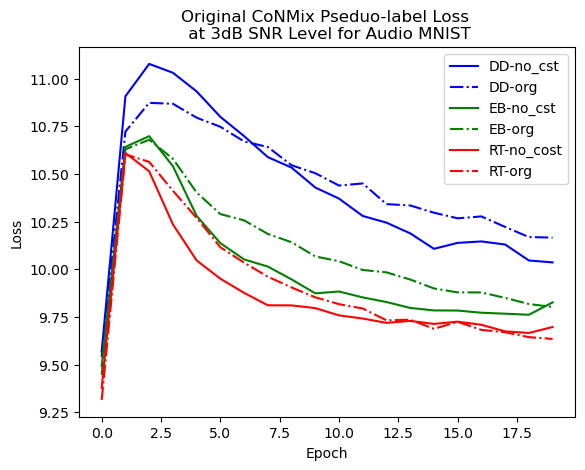}}
    \hfill
    \subfloat[\label{fig:PL-3dB-AM}]{\includegraphics[width=0.33\linewidth]{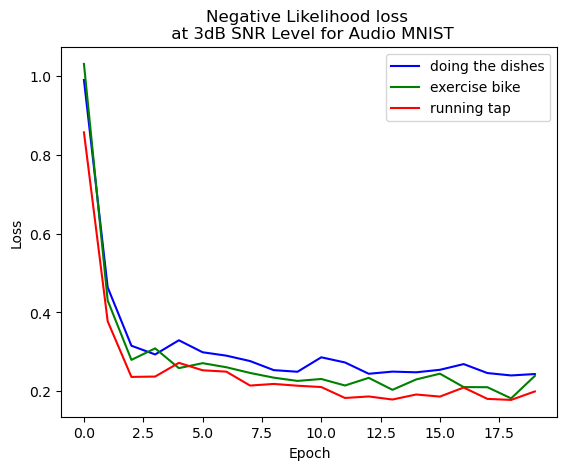}}
    \caption{The accuracy results at 10 dB ((a), (b), (c), and (d)) and 3 dB ((g), (h), (i), and (j)) SNR level for AM. To illustrate, upd (replace Pseudio-label loss (\ref{CoNMix_pl_ce_loss}) with NLL loss (\ref{CoNMix_pl_nll_loss})) (see \ref{sec:methodology})), org (original CoNMix Solution), no\_cst (exclude consistency loss (\ref{eq:L_cons})), no\_pl (exclude pseudo-label loss (\ref{CoNMix_pl_ce_loss})), and no\_nm (exclude Nuclear-norm Maximization loss (\ref{Nuclear_norm_loss})). CoNMix Pseudo-label Loss under three 10 dB ((e) and (f)) and 3 dB ((k) and (l)) SNR level background noises, such as DD ('doing the dishes'), EB ('exercise bike'), and RT ('running tap'), for AM.}
    \label{fig:allResults-PA-10dB-AM}
\end{figure*}

Even though 'org' does not present a negative adaptation effect in Figure.\ref{fig:PA-10dB-DD-AM},~\ref{fig:PA-10dB-EB-AM},~\ref{fig:PA-10dB-RT-AM},~\ref{fig:PA-3dB-DD-AM},~\ref{fig:PA-3dB-EB-AM}, and~\ref{fig:PA-3dB-RT-AM}, 'no\_pl' is the best solution among 'org', 'upd', 'no\_pl', and 'no\_cst'.

The original CoNMix pseudo-label loss (Eq.\ref{CoNMix_pl_ce_loss}) has an increased tendency (see Fig.\ref{fig:PL-10dB-org-AM} and~\ref{fig:PL-3dB-org-AM}). After replacing (Eq.\ref{CoNMix_pl_ce_loss}) with NLL loss (Eq.\ref{CoNMix_pl_nll_loss}), the pseudo-label loss is convergent (see Fig.\ref{fig:PL-10dB-AM} and~\ref{fig:PL-3dB-AM}).
\subsection{Summary}
It is worth noting that for CoNMix, the pseudo-label loss function (see Eq.\ref{CoNMix_pl_ce_loss} and Eq.\ref{CoNMix_pl_nll_loss}) contributed in a limited way to test-time adaptation performance. To tackle the issue in a post-mortem manner, we retrospectively manipulated the scaling factor $\lambda_2$ of the pseudo-label loss function. Specifically, for the results shown in Table~\ref{tab:resultsAll} section~\ref{sec:evaluation}, we removed the pseudo-label loss entirely for the experiment on AM under 3.0 dB and 10.0 dB SNR levels in Eq.\ref{STDA_loss}.

\end{document}